%% file: egpaper.tex
\documentclass[10pt,twocolumn,letterpaper]{article}

\usepackage{iccv}
\usepackage{times}
\usepackage{epsfig}
\usepackage{graphicx}
\usepackage{amsmath}
\usepackage{amssymb}

\newcommand{\negneg}{\mathrel{\hspace*{-3pt}\mathord{-}\hspace*{0pt}\mathord{-}}}

\usepackage{boldline}
\usepackage{multirow}
\usepackage{multirow,diagbox,tabularx,blindtext}
\usepackage{bbm}

\usepackage{adjustbox}

\usepackage{diagbox}

\usepackage{enumitem}

\usepackage{mathtools}

\usepackage[pagebackref=true,breaklinks=true,letterpaper=true,colorlinks,bookmarks=false]{hyperref}

\iccvfinalcopy % *** Uncomment this line for the final submission

\ificcvfinal\fi

\usepackage[dvipsnames]{xcolor}

\usepackage{algorithm}
\usepackage[noend]{algpseudocode}
\makeatletter
\def\BState{\State\hskip-\ALG@thistlm}
\makeatother

\algnewcommand\algorithmicforeach{\textbf{for each}}
\algdef{S}[FOR]{ForEach}[1]{\algorithmicforeach\ #1\ \algorithmicdo}

\renewcommand\algorithmicthen{}

\algnewcommand{\IfThenElse}[3]{% \IfThenElse{<if>}{<then>}{<else>}
  \State \algorithmicif\ #1\ \algorithmicthen\ #2\ \algorithmicelse\ #3}

\algnewcommand{\IfThenOneLine}[1]{% \IfThenElse{<if>}{<then>}{<else>}
  \State \algorithmicif\ #1\ \algorithmicthen\ }

\usepackage{tabularx}
\usepackage{booktabs}% http://ctan.org/pkg/booktabs
\newcolumntype{R}{>{\centering\arraybackslash}X}

\algrenewcommand\algorithmicindent{0.5em}%

\newlength\algowd

\usepackage{eqparbox}

\usepackage{adjustbox}
\newcolumntype{Y}{>{\centering\arraybackslash}X}

\newcolumntype{?}{!{\vrule width 1.5pt}}
\usepackage{multirow}

\usepackage{subfig}

\usepackage{longtable}

\usepackage{tabularx}
\newcolumntype{L}{>{\raggedright\arraybackslash}X}

\usepackage{amsmath,epsfig}

\usepackage{amssymb}

\usepackage{rotating}

\begin{document}

%%%%%%%%% TITLE
\title{High Performance Convolution Using Sparsity and Patterns for Inference in Deep Convolutional Neural Networks}
% \title{Efficient Convolution Using Sparsity and Patterns for Inference in Deep Convolutional Neural Networks}

% \author{Hossam Amer, Ahmed H. Salamah, 
% \\
% \\
% Paper ID ****
% }

% \author{\textbf{Hossam Amer$^1$}
% \and
% \textbf{Ahmed H. Salamah$^1$}
% \and
% \textbf{Ahmad Sajedi$^2$}
% \and
% \textbf{En-hui Yang$^1$}\and 
% $^1$University of Waterloo \hspace{4cm}
% $^2$University of Toronto\\
%  {\tt \small\{hossam.amer,ahamsalamah,ehyang\}@uwaterloo.ca, ahmad.sajedi@mail.utoronto.ca}\\
% %{\tt ahmad.sajedi@mail.utoronto.ca}
% }

\author{\textbf{Hossam Amer$^*$}
\and
\textbf{Ahmed H. Salamah$^*$}
\and
\textbf{Ahmad Sajedi$^{**}$}
\and
\textbf{En-hui Yang$^*$}\and 
$^*$University of Waterloo \hspace{4cm}
$^{**}$University of Toronto\\
 {\tt \small\{hossam.amer,ahamsalamah,ehyang\}@uwaterloo.ca, ahmad.sajedi@mail.utoronto.ca}\\
%{\tt ahmad.sajedi@mail.utoronto.ca}
}

\maketitle
\ificcvfinal\thispagestyle{empty}\fi

%%%%%%%%% ABSTRACT
\begin{abstract}
   Deploying deep Convolutional Neural Networks (CNNs) is impacted by their memory footprint and speed requirements, which mainly come from convolution. Widely-used convolution algorithms, im2col and MEC, produce a lowered matrix from an activation map by redundantly storing the map's elements included at horizontal and/or vertical kernel overlappings without considering the sparsity of the map. Using the sparsity of the map, this paper proposes two new convolution algorithms dubbed Compressed Pattern Overlap (CPO) and Compressed Pattern Sets (CPS) that simultaneously decrease the memory footprint and increase the inference speed while preserving the accuracy. CPO recognizes non-zero elements (NZEs) at horizontal and vertical overlappings in the activation maps. CPS further improves the memory savings of CPO by compressing the index positions of neighboring NZEs. In both algorithms, channels/regions of the activation maps with all zeros are skipped. Then, CPO/CPS performs convolution via Sparse Matrix-Vector Multiplication (SpMv) done on their sparse representations. Experimental results conducted on CPUs show that average per-layer time savings reach up to 63\% and Compression Ratio (CR) up to 26x with respect to im2col. In some layers, our average per layer CPO/CPS time savings are better by 28\% and CR is better by 9.2x than the parallel implementation of MEC. For a given CNN's inference, we offline select for each convolution layer the best convolutional algorithm in terms of time between either CPO or CPS and im2col. Our algorithms were selected up to 56\% of the non-pointwise convolutional layers. Our offline selections yield CNN inference time savings up to 9\% and CR up to 10x. 
\end{abstract}

%%%%%%%%% BODY TEXT
\section{Introduction}

Deep Convolutional Neural Networks (CNNs) have become the defacto approach for the majority of computer vision tasks. For example, deep CNNs on imageNet classification dataset have produced an accuracy of 90.2\% \cite{pham2020meta}. Deep CNNs encompass a cascaded and large number of layers that extract features from the input image without any domain knowledge. As the number of these layers continues to grow, CNNs become deeper with increasing memory and computational requirements. These requirements pose certain challenges in deploying deep CNNs in end devices such as mobile phones or IoT devices \cite{goodfellow2016deep}.

Among these many layers are the convolutional layers, which occupy 70-90\% of CNN's computational requirements \cite{hadjis2015caffe}. Currently employed convolution algorithms utilize GEneral Matrix to Matrix Multiplication (GEMM) method, where activation maps are transformed into a 2D lowered matrix that is multiplied by a kernel to 
produce the output feature maps \cite{chellapilla2006high_im2col, hadjis2015caffe, cho2017mec}.

Lowering-based approaches with GEMM, especially im2col, are widely-used in current Machine Learning (ML) frameworks due to their efficiency, but they did not fully consider the fact that Rectifier Linear Unit (ReLU) is the typical activation function in CNNs \cite{chellapilla2006high_im2col, hadjis2015caffe}. Using ReLU leads to sparse activation maps. Experiments carried out on popular CNNs in classification task reveal that 30-50\% of convolutional (conv) layers have a density (i.e., portion of Non-Zero Elements (NZEs)) of activation maps between 0.05 and 0.5 \cite{szegedy2016inceptionv3, he2016identityresnet}. The lowering-based approaches operate on both zero and NZEs and store them in the lowered matrix, while NZEs are only contributing to the output feature maps. On top of that, both zero elements and NZEs in activation maps included at horizontal and/or vertical kernel overlapping patterns are repetitively stored in the lowered matrix, which create some redundancy. In addition, NZEs in activation maps normally appear near each other, so knowing one NZE index and the index pattern of its neighbors can help reconstruct the rest. Moreover, some channels or regions in the activation maps are entirely zero and can be skipped. These insights lead to the question of whether it is possible to utilize activation map's sparsity and patterns to develop fast and memory-efficient algorithms for computing convolution in CNNs.

This paper proposes two convolution algorithms dubbed Compressed Pattern Overlap (CPO) and Compressed Pattern Sets (CPS) implemented on CPUs that simultaneously decrease the memory footprint and increase the inference speed while preserving the accuracy of deep CNNs. CPO recognizes NZEs at horizontal and vertical overlappings in the activation maps. CPS further improves the memory footprint savings upon CPO by compressing neighboring NZE index positions. Both CPO and CPS also include Skip flags for channels or regions in activation maps that contain no NZEs. After activation map compression, CPO/CPS performs convolution using Sparse Matrix Vector Multiplication (SpMv), where the output resulting from each channel and kernel is performed independently \cite{bell2008efficient}.

To apply CPO/CPS to the whole inference process of a given CNN, we observe that the average density at each conv layer is approximately stationary across images. Given a target CNN for inference and a small set of images, we offline select a better convolution method between im2col and our algorithms in terms of time. This results in a hybrid convolution selection method. When applied to the end-to-end inference on the given CNN, the hybrid convolution selection method shows maximum end-to-end inference time savings of 9\%, while the average gain in Compression Ratio (CR) across layers go up to 10x. Our contributions are:

\begin{itemize}[noitemsep,nolistsep]
    \item We propose CPO and CPS based on sparsity and patterns of the activation maps with per-layer time savings up to 63\%, CR up to 26x with respect to im2col, while maintaining accuracy. (Section \ref{sec:motivation}-\ref{sec:hybrid}, \ref{sec:experimental_results})
    \item To the best of our knowledge, this is the first paper that showed both time and space savings obtained from its convolution algorithms in real inference. (Section \ref{sec:related_work}, \ref{sec:experimental_results})
    \item The proposed hybrid convolution selection method provides inference time savings up to 9\% and average CR across layers up to 10x. (Section \ref{sec:hybrid}, \ref{sec:experimental_results})
\end{itemize}

%%%%%%%%%%%%%%%%%%%%%%%%%%%%%%%
\section{Related Work}
\label{sec:related_work}

% Group1: HW accelerators for sparsity
% Group2: Pruning and lossy stuff… weight focus
% Group3: Constructing on sparsity of the feature maps… very very few papers
% speed up sparsity is only 0.9

A wide line of research has been dedicated to compressing and accelerating CNNs by considering sparsity for speed and memory benefits \cite{chetlur2014cudnn, hadjis2015caffe, wen2016learning, cho2017mec, chen2018escoin, hackel2018inference, rhu2018compressing, shi2020communication, evans2020jpeg, hoefler2021sparsity}. This line of research can be divided into three main classes. The first class is considering sparsity with focus on dedicated accelerators that use sparsity to overcome its memory challenges \cite{cong2014minimizing, chen2016eyeriss, albericio2016cnvlutin, han2016eie, parashar2017scnn, chen2019eyeriss}. In this paper, we review a subset of papers for the other two classes.

\textbf{Sparsity of Kernels:} Direct sparse convolution in \cite{park2016holistic, park2016faster} prunes the unnecessary kernel weights in CNNs and stores the non-zero weights in a Compressed Sparse Row (CSR) format in order to  perform convolutions while maintaining the accuracy of the CNN classifiers. Furthermore, a three-stage pipeline called deep compression was successfully proposed in \cite{han2015deep} where weights are pruned and stored in CSR format, quantized, and then huffman encoded to reduce the computational requirements of CNNs, but with a drop of accuracy up to 2.6\% in some CNNs. Method in \cite{liu2015sparse} proposed a decomposition of weights where their sparsity is encouraged. Then,
convolutions are executed via sparse matrix multiplications during inference time. The papers \cite{changpinyo2017power, guo2016dynamic, li2016pruning, lebedev2016fast, lecun1989optimal, hassibi1993optimal, kundu2020pre} also focused on kernel sparsity.

\textbf{Sparsity of Activation Maps:} Few papers considered the sparsity of input activation maps although they can play a role in optimizing CNNs due to its relatively larger size than kernel weights. For instance, the approach in \cite{georgiadis2019accelerating} successfully induced sparsity to the input maps by retraining using the L1 loss. Then, inputs are quantized, and entropy coded to reduce memory bandwidth. However, the reported speed-up is not in terms of the actual inference time, but rather in terms of the reduction of NZEs in input maps.
The method in \cite{dong2019exploiting} explored the sparsity of input maps to accelerate inference and was applied to Lidar-based detection or character recognition with reasonable gains. As stated in \cite{dong2019exploiting}, this method works well in certain tasks or models with a limited and high range of sparsity.
Besides, authors in \cite{kurtz2020inducing} induced $\approx$10\% more sparsity on average in CNNs by introducing a threshold-based ReLU activation function and parallelized sparse convolution with noticeable gains in terms of time. However,
they did their tests in a sparsity-induced CNN architecture without reporting actual space savings. Furthermore, the method in \cite{shi2017speeding} encoded the input maps via the Coordinate (COO) lists representation and performed convolutions on specific conv layers with density only less than 0.1 without retraining and drop in accuracy. In \cite{fan2019cscc}, speed and memory gains over a set of layers were achieved via a CSR-based convolution without considering the horizontal kernel overlapping patterns.

This paper differs from the aforementioned work in three distinct ways: (1) our work considers the sparsity of activation maps rather than kernels. (2) CPO/CPS considers sparsity and patterns such as avoiding repetitive storage of NZEs at horizontal and vertical overlappings, reducing the storage of NZE indices, and skipping all-zero channels or regions. (3) CPO/CPS does not require altering CNNs, do not cause any drops in classification accuracy, deployable in ML frameworks, and provide per-layer CR gains up to 26x and end-to-end inference time savings up to 9\%. 

\section{High Performance Convolution Algorithms}
\label{sec:method}

\subsection{Sparsity and Patterns Motivation}
\label{sec:motivation}

As defined in Table \ref{tab:conv_notations}, consider $n$ activation maps with $I_c$ channels. The resulting output with dimensions $I_n \times O_h \times O_w \times K$ from the convolution operation with a kernel W is calculated by:
\begin{flalign}
 & O(n, y, x, k)  = \nonumber \\   \sum_{c = 0}^{I_{c}-1} & \sum_{h = 0}^{K_{h}-1}  \sum_{w = 0}^{K_{w}-1}  I(n, c, y+h, x+w)  W(h, w, c,  k)
\label{eq:conv_definition}
\end{flalign}
\noindent where the output spatial dimensions are $O_{h, w}$ = 1 + $\frac{I_{h,w} - K_{h,w}}{s_{h,w}}$. Equation \ref{eq:conv_definition} shows that the output feature map at each location $(y,x)$ with a fixed set of kernel weights $W$ is induced by a series of partial sums and multiplications between the input, $I$, and the kernel, $W$. Normally, the output is computed using matrix multiplications between a lowered matrix of $I$ and $W$. As $I$ in typical CNNs is an output of ReLU activation function defined as $max(0, z)$, it is typically sparse. Zeros in activation maps do not contribute to the output feature maps. Thus, we can reduce the redundancy of activation maps and avoid unnecessary zero computations leading to time/memory savings. 

Let us assume a 4x4 kernel with strides 1 and one multi-channel 8x8 input activation map. Lowering-based approaches such as \cite{chellapilla2006high_im2col, hadjis2015caffe} and \cite{cho2017mec} convert the activation map into a 2D lowered matrix to perform fast matrix multiplication with the kernel. For instance, the default convolution algorithm in the majority of ML frameworks utilizes GEMM using im2col, where inputs are first transformed into a lowered matrix with redundancy. Im2col's lowered matrix is extracted from patches of the kernel size ($4x4$), where each patch in each channel is inserted in a separate row of the lowered matrix. New patches are created by shifting the kernel to either right or bottom by a stride of 1. Im2col's lowered matrix will be relatively large because it repetitively stores a fixed input data value at multiple horizontal and vertical kernel overlappings when this value contributes to the different index positions of the output. 

\begin{figure*}[!ht]
  \centering
  \vspace{-0.8em}
  \setlength{\belowcaptionskip}{-3ex}
  \includegraphics[width=0.9\textwidth,height=0.38\textwidth]{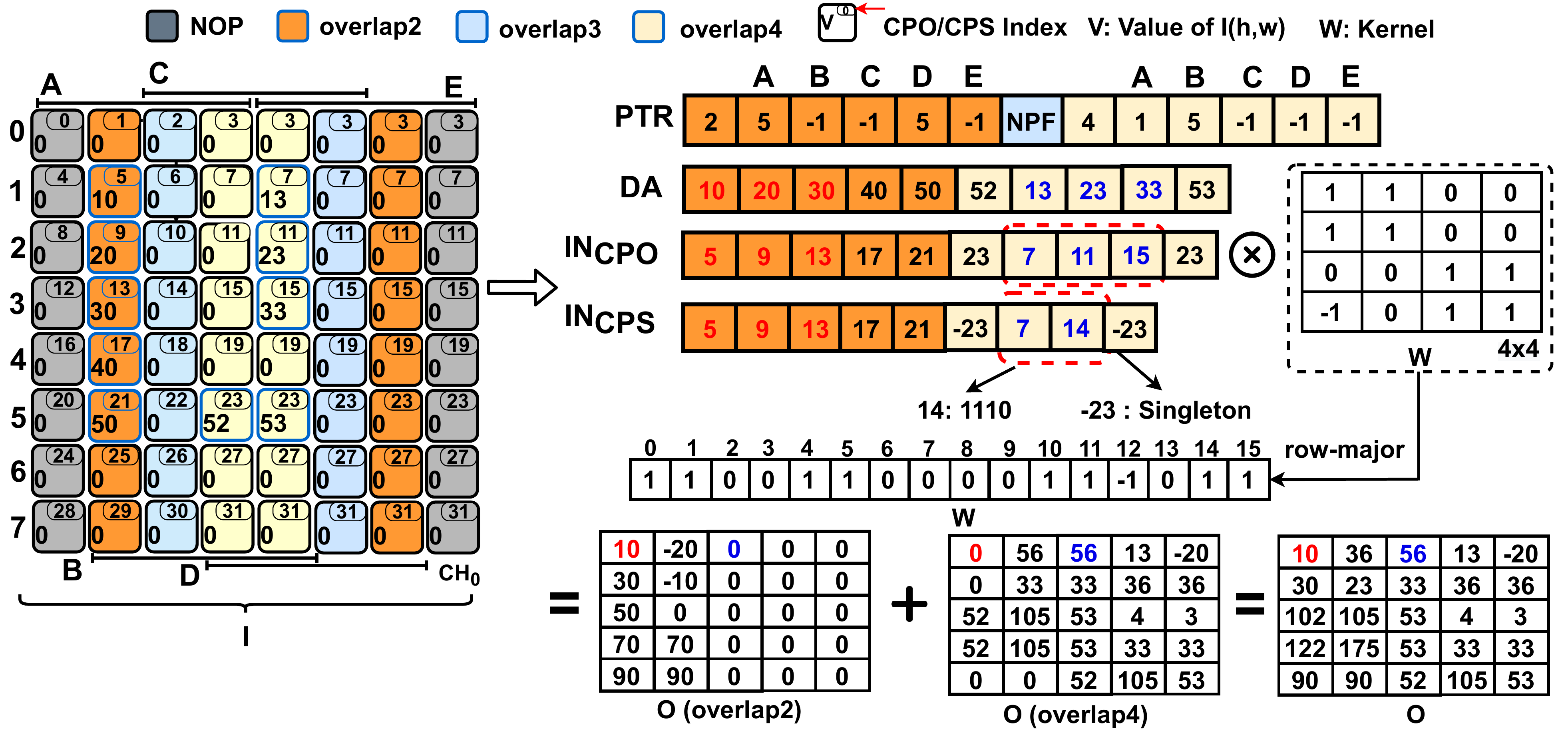}
%   \caption{\textit{Left:} Example of input activation map with $I_h = I_w = 8$, $K_h = K_w = 4$, $s_h = s_w = 1$, $I_n = I_c = K = 1$. \textit{Middle:} A submatrix from MEC's lowered matrix constructed from row 3 of $I$. \textit{Right:} CPO/CPS Encoding and Convolution.}
  \caption{\textit{Left:} Example of input activation map with $I_h = I_w = 8$, $K_h = K_w = 4$, $s_h = s_w = 1$, $I_n = I_c = K = 1$. \textit{Right:} CPO/CPS Encoding and Convolution. Red and blue fonts in $IN$ and $DA$ correspond to their contributions in $O$.}
  \label{fig:org_feature_map_lowered_mat}
\end{figure*}

MEC divides $I$ into $O_{w}$ horizontal partitions with the size of $I_{h} \times K_{w}$ where each partition is separated by $s_{w}$. For example in the Figure \ref{fig:org_feature_map_lowered_mat} we have $5$ horizontal partitions A, B, C, D, and E with the size of $8 \times 4$ (e.g., Partition A is $I[0:7, 0:3]$). Each partition is then converted into a row in the lowered matrix. Hence, elements of $I$ included at horizontal kernel overlapping sites are repeatedly stored at multiple rows of the lowered matrix. For example, the first orange column in $I$ is an overlapping region that overlap horizontally between A, B (overlap2 type). The first blue column has a horizontal overlap between A, B, C (overlap3 type), and the first yellow column has a horizontal overlap between A, B, C, D (overlap4 type). Thus, values of the orange, blue, and yellow columns appear 2, 3, and 4 times, respectively, in the lowered matrix with fixed patterns. 

From this discussion, we are motivated to perform convolution via other representations that consider sparsity of $I$ and both the horizontal and vertical patterns. Section \ref{sec:alg_overview} provides an overview on the components used in CPO/CPS.

\begin{table}[htbp]
    \centering
    \caption{Mathematical Notations for Convolution.} \hspace{2cm}
\resizebox{0.84\columnwidth}{!}{%   
\begin{tabular}{?c?c?c?}
\specialrule{.1em}{.05em}{.05em}  \hline
   \textbf{Symbol} & \textbf{Notation} & \textbf{Spatial Dimensions} \\
  \specialrule{.1em}{.05em}{.05em}  \hline
        $I$ & Input & $I_n \times I_h \times I_w \times I_c$\\
        \specialrule{.1em}{.05em}{.05em}  \hline
        $O$ & Output & $I_n \times O_h \times O_w \times K $\\
        \specialrule{.1em}{.05em}{.05em}  \hline
        $W$ & Kernel & $K_h \times K_w \times I_c  \times K$\\
        \specialrule{.1em}{.05em}{.05em}  \hline
        $s_w, s_h$ & Kernel stride & \\
     \specialrule{.1em}{.05em}{.05em}  \hline
\end{tabular}
}
\label{tab:conv_notations}
\vspace{-1em}
\end{table}

\subsection{Overview of CPO and CPS}
\label{sec:alg_overview}

%%%%%%%%%%%% 
% 1) INDEXING
% 2) NZE representations 
% 3) CPS IN difference
% 4) conv 

CPO and CPS first split $I$ into $O_{w}$ horizontal partitions separated by $s_w$. For each cell in the first partition of $I$, CPO/CPS assigns a row-major index from 0 to $I_{h}*K_{w} - 1$. In partition A of figure \ref{fig:org_feature_map_lowered_mat}, we assign indices from 0 to 3 for the first row, 4 to 7 for the second row, until 28 to 31 for the last row. In the other partitions, we only assign indices to columns which appear in the new horizontal shift of the kernel window. As a result, we only assign indices in column 4 at partition B. At column 4, we assign the index 3 for the cell at row 0, index 7 for the cell at row 1, and finally the index 31 for the cell at row 7. Partitions C to E follow the exact indexing methodology as partition B. 

For CPO and CPS convolution, analysis is done on the nature of conv layers in 6 CNNs used in image classification. Results from the experiment show that 58.5\% on average of the convolution layers utilize pointwise convolution. On average, $37.3 \%$ of the layers has $s_h = s_w = 1$ and no pointwise convolution. We focus CPO/CPS on non-pointwise kernels whose strides are $s_h = s_w = 1$, but we examine the impact of CPO/CPS on real CNN inference.

% NZE and sparsity
To eliminate zero multiplications in convolution, CPO/CPS considers NZEs of $I$, the horizontal and vertical kernel overlappings. Inspired by CSR, CPO/CPS introduces structures to store the NZEs of $I$: pointer $ptr$, data $DA$, and index $IN$. For each channel, creation of $ptr$, $DA$, and $IN$ is done first for the Non-Overlapping (NOP) regions and then overlap2, overlap3, etc., and  overlap$K_w$ regions. To perform convolution, CPO/CPS uses the SpMv between their encoding representations and the kernel. In Section \ref{sec:cpo} and \ref{sec:cps}, we provide a detailed explanation for the $ptr$, $DA$, and $IN$ creation and their usage to perform convolution. We assume that $I_n=1, K=1$, there are several overlap regions in $I$; other cases are  straightforward extensions and will be provided in the Supplementary Information (SI). Also, space complexity derivations for CPO and CPS can be found in the SI.

% In addition, we assume that $K=1$ in convolution to simplify the presentation.

%%%%%%%%%%%%%%%%%%%%% Up there is the luck

\subsection{CPO Encoding and Convolution Algorithms}
\label{sec:cpo}

% Ordering first
% we encode 
% we convolve using SpMv
% Implementation details: Improves Cache locality + Loop Unrolling + Continuous memory access for encoding output + …. ?
% zero skipped

% To illustrate the creation of $ptr$, $DA$, and $IN$ for CPO, let us look at $I$ and corresponding CPO encoding output in Figure \ref{fig:org_feature_map_lowered_mat}. 

Given $I$ and its corresponding indices and partitions obtained in Section \ref{sec:alg_overview}, we create $ptr$, $DA$, and $IN$ for CPO. Depending on the amount of padding ($Pad\_Left$) on the left of $I$, CPO skips regions with zero density and starts processing from the next region containing NZEs. In Figure \ref{fig:org_feature_map_lowered_mat}, no information for NOP is inserted in all structures because the first and last column in $I$ are all zeros.

The first value of $ptr$ indicates the overlapping type (e.g., 2 for overlap2 type) while the second value represents the number of NZEs in the first column of the given overlap type (e.g., the first orange column of partition A for overlap2). Then, we insert the values and indices of NZEs in $DA$ and $IN_{cpo}$, respectively (e.g., (10, 20, 30, 40, 50) in $DA$ and indices (5, 9, 13, 17, 21) in $IN_{cpo}$). If the next partitions do not have any new columns in a given overlap type, we insert -1 in $ptr$ and nothing for both $DA$ and $IN_{cpo}$; otherwise, we write the cumulative number of NZEs of the given overlap type in $ptr$ and their corresponding NZEs' data values and indices in $DA$ and $IN_{cpo}$, respectively. In overlap2 for example, the corresponding $ptr$ of partitions B, C, and E are -1. Partition D has 5 in $ptr$ to indicate a new column in overlap2 type, but with no cumulative change in terms of the number of NZEs with respect to partition A. This process repeats for all overlap types which have NZEs (see overlap4 type in Figure \ref{fig:org_feature_map_lowered_mat}). For more time/space savings, we insert skip flags in $ptr$ for channels/regions that have only zero elements. No Ptr Flag (NPF) is for overlap regions, and No Ptr Channel (NPC) is for channels. 

Algorithm \ref{alg:cpoEncodeAlg} outlines the CPO encoding that consists of three parts. In all three parts, G function inserts the data and corresponding index of NZEs of the given column $w$ in $DA$ and $IN$ vectors, respectively. This function uses nz\_ptr to obtain cumulative number of NZEs in the given overlap region. First, lines 11-16 are for NOP, which holds NZEs in activation maps contributing only to the vertical positions of the output (i.e, first and last column in $I$). Second, lines 17-24 are for overlapping pointers starting from pType until $K_w-1$ \footnote{pType = 0 is for NOP, pType = 1 is for overlap2, etc. and pType = $K_w-1$ is for overlap$K_w$}. Third, lines 27-29 are for the overlap$K_w$ region which includes NZEs that contribute $K_w$ times horizontally in the convolution (e.g. yellow columns of overlap4 in Figure \ref{fig:org_feature_map_lowered_mat}). In lines 31-37, we determine NPF or NPC flags. 

\begin{algorithm}
    \caption{CPO Encoding Algorithm}
    \label{alg:cpoEncodeAlg}
    \begin{algorithmic}[1]
        \Function{G}{$I$, $w$, $Z$, $ptr$, $DA$, $IN$, nz\_ptr, ptr\_sh}
            \For{$h = 0$ to $I_h$}
                \IfThenOneLine{$I(h, w) \neq 0$ \hspace{0.1em} \textrm{\textbf{then}}} {$DA$.insert($I(h,w)$), 
                \State $IN$.insert($Z + (h * K_w)$), nz\_ptr++}
            \EndFor
            % \State \Return $ptr$, $DA$, $IN$, $nz\_ptr$, $ptr\_sh$, $i$
        \EndFunction
        \Procedure{cpoEncode}{}
            % \State $start = 0,$ $end = 8,$
            % \State Compute $\varepsilon_k$ from, where $start \leq k \leq end/2$
            %  \State Based on, $\forall_{j \in [end+1, end+4]}$ $\varepsilon_{j} = \varepsilon_{j-8},$
            %  \State $start = start + 4,$ 
            %  \State $end = end + 4,$ go to $\#3$
            \State \textbf{Input:} $ I$, $K_w$, $Pad\_Left$
            \State \textbf{Output:} $ptr$, $DA$, $IN$
            % \State $ptr\_sh=0$,\hspace{0.1em}$nz\_ch=0 $,\hspace{0.1em}$val\_sh=0$,\hspace{0.1em}$l=Padding_{Left}$ ,\hspace{0.1em}$i=0$ % initiate zero to all vairables
            % \State Initialize $ptr\_sh, nz\_ch, val\_sh, i$ to 0, pType to Pad\_Left
            \State Initialize ptr\_sh, val\_sh to 0
            \For{each channel \textbf{in} $I_{c}$} 
                \State nz\_ch, nz\_ptr = 0, pType = Pad\_Left
                \If{pType = 0 \textrm{\textbf{then}}} %\Comment{If VALID Padding}
                    \State nz\_ptr = 0, $ptr$[ptr\_sh++] = 0, val\_sh = 2
                    % \For{ \{$w, \hat{w}$\} \textbf{in} $[\{ 0, 0\},\{ I_w-1, K_w-1\}]$}
                    \For{ \{$w, \hat{w}$\} \textbf{in} $\{\{ 0, 0\},\{ I_w-1, K_w-1\}\}$}
                    % \For{ $ w , \hat{w} \in \{\{ 0\},\{ 0\},\{ \{I_w-1\}, \{K_w-1\}\}$} 
                        \State G($I$, $w$ , $\hat{w}$, $ptr$, $DA$, $IN$, nz\_ptr, ptr\_sh)
                        \State $ptr$[ptr\_sh++] = nz\_ptr 
                    \EndFor
                    \State nz\_ch += nz\_ptr, \hspace{0.1em} pType++
                \EndIf
                \State $\check{w} = I_w - ($pType$ + 1)$,\hspace{0.1em} idx = $O_w$ - pType
                \For{$w$ = pType to $K_w$ - 1} %\Comment{After Padding Col.}
                    \State nz\_ptr = 0 , $ptr$[ptr\_sh++] = pType
                        \For{ \{$\bar{w}, \hat{w}$\} \textbf{in} $\{\{ w, w \},\{ \check{w}, \check{w}-($idx$-1) \}\}$}
                    % \For{ $\bar{h}, \hat{h}$ \textbf{in} $\{h, \check{h}\}, \{ h,\check{h}+(idx-1)\}$}
                        % \For{$h = 0$ to $I_h$}
                        %     \IfThenOneLine{$I(w,\bar{h}) \neq 0$ \hspace{0.1em} \textrm{\textbf{then}}} {$DA[i] = I(w,\bar{h})$, 
                        %     \State $IN[i] = \hat{h} + (w * Kw)$ , \hspace{0.1em} $i\plusplus$ , \hspace{0.1em}$nz\_ptr\plusplus$}
                        %     \State $ptr[ptr\_sh\plusplus] = nz\_ptr $
                        % \EndFor
                            \State G($I, \bar{w}$, $\hat{w}$, $ptr$, $DA$, $IN$, nz\_ptr, ptr\_sh)
                        
                    \EndFor
                    % \State $nz\_ch \plusequals nz\_ptr$, \hspace{0.1em} $\check{h}\negneg$, \hspace{0.1em} $pType\plusplus$, \hspace{0.1em} $idx\negneg$
                    % \If{$nz\_ptr > 0$ \hspace{0.1em} \textrm{\textbf{then}}} $ptr\_sh \plusequals idx$ , 
                    % \State $ptr[ptr\_sh]=nz\_ptr$, $ptr\_sh \plusequals (O_w+1) - idx $
                    % \Else { $ptr[ptr\_sh\plusplus] =$ NPF}  
                    % \EndIf
                    \If{nz\_ptr $>$ 0 \textrm{\textbf{then}}}  
                    \State $ptr$[ptr\_sh + idx] = nz\_ptr, ptr\_sh += $(O_w+1)$
                    \Else { $ptr$[ptr\_sh - 1] = NPF} \Comment{If Skip pointer}  
                    \EndIf
                    \State nz\_ch += nz\_ptr , pType++, idx++ , $\check{w} \negneg$
                \EndFor % {$w = pType$ to $K_w - 1$ to }
                \State idx = 0, $ptr$[ptr\_sh++] = pType 
                \For{$w = K_w - 1$ to $I_w - K_w$}
                %     \For{$h = 0$ to $I_h$}
                %         \IfThenOneLine{$I(w,h) \neq 0$ \hspace{0.1em} \textrm{\textbf{then}}} {$DA[i] = I(w,h)$, 
                %         \State $IN[i] = w - (idx - 1) + (h * Kw)$,\hspace{0.1em} $i\plusplus$},\hspace{0.1em} $nz\_ptr\plusplus $
                %   \EndFor 
                    \State G($I$, $w$, $w$ - idx, $ptr$, $DA$, $IN$, nz\_ptr, ptr\_sh)
                    \State $ptr$[ptr\_sh++] = nz\_ptr
                    %  \State $ptr[ptr\_sh\plusplus] = nz\_ptr$,\hspace{0.1em} 
                    , idx++
                \EndFor
                 \State nz\_ch += nz\_ptr 
                %  \IfThenOneLine{$nz\_ch = 0$ \hspace{0.1em} \textrm{\textbf{then}}}{$ptr\_sh -= (K_w - 1) - ptr\_start + val\_sh$,$ptr[ptr\_sh] = NPC,$ ptr[ptr\_sh + 1:O_w + val\_sh +1] = -1$}
                 \If {nz\_ch=0 \textrm{\textbf{then}}} \Comment{If Skip Channel}
                     \State ptr\_sh -= ($K_w$-1) + val\_sh - Pad\_Left, 
                    %  \State $ptr[ptr\_sh\plusplus]$ = NPC, 
                     \State $ptr$[ptr\_sh++] = NPC, 
                     \State $ ptr$[ptr\_sh : ($O_w$ + val\_sh +1)] = -1
                    %   \For{$i = 1$ to $i< O_w + val\_sh +1$}
                    %         \State $ptr[ptr\_sh + i] = -1$
                    %     \EndFor 
                 \ElsIf{nz\_ptr $>$ 0  \textrm{\textbf{then}}}
                        ptr\_sh += $O_w$ + 1
                        % \For{$i = 1$ to $i< O_w + val\_sh +1$} 
                        %     \State $ ptr[ptr\_idx + i] = -1$
                        % \EndFor
                \Else {\hspace{0.1em} $ptr$[ptr\_sh++] = NPF}
                    \State $ptr$[ptr\_sh : ($O_w$ + val\_sh + 1)] = -1
                 \EndIf
            \EndFor % Channel Loop
        \EndProcedure
    \end{algorithmic}
    
\end{algorithm}

Algorithm \ref{alg:cpoConvAlg} performs convolution between the CPO encoding of $I$ and kernel that is done independently for each channel and kernel. Algorithm \ref{alg:cpoConvAlg} decomposes the final output into a summation of partial outputs in each overlap type. We show the partial output from overlap2 (O(overlap2)) and overlap4 (O(overlap4)) in Figure \ref{fig:org_feature_map_lowered_mat}, where the final output is the cumulative sum between these partial outputs. The procedure is explained as follows. The kernel is transformed to a row-major vector (see Figure \ref{fig:org_feature_map_lowered_mat}). If NOP exists, then we compute the SpMv on the first partition, 0, and the last partition, $O_w$-1. In Figure \ref{fig:org_feature_map_lowered_mat}, NOP does not have NZEs and we skip it (line 15). The rest of the NZEs are processed in the increasing order of the overlap type. In the convSpMv function between lines 1 to 8, each NZE contributes at most $K_h*pType$ to the output, where pType is the overlap type. For example, the first NZE in overlap2 is 10 in $DA$ contributes to four output indices (i.e, O[0:1][0:1]). Given the $IN_{CPO}[0]$ corresponding to 10, we compute the two co-ordinates, y and s, of the output and kernel position (t) based on lines 3 and 4 in Algorithm \ref{alg:cpoConvAlg}. Except for overlap3 that is skipped by NPF, the rest of the NZEs in overlap2 type follows the steps for 10, and elements in  overlap4 follows the process of overlap2, but with pType=3. Indexing of $ptr$, $IN$, and $DA$ in Algorithm \ref{alg:cpoConvAlg} relative to each channel is omitted for simplicity, i.e, line 14 reads the first value of ptr in each channel.

\begin{algorithm}
    \caption{CPO Convolution Algorithm}
    \label{alg:cpoConvAlg}
    \begin{algorithmic}[1]
      \Function{convSpMv}{$K_h$, $K_w$, $O_h$, pType, x, s, index}
            \For{$l = 0$ to $K_h$}
                \State y = (index / $K_w$) - l
                \State t = (index \% $K_w$ + $l*K_w$)
                \If{y $\ge$ 0 \hspace{0.1em} $\&$  y $<$ $O_h$}         \State O[y][s] += DA[x]*W[t]   
                   % the rest of the contributions
                   \For{$i = 1$ to $pType$} 
                     \State O[y][s+i] += DA[x]*W[t-i]
                   \EndFor % i loop
               \EndIf % End if loop
             \EndFor % l loop
    \EndFunction
        \Procedure{cpoAlg}{}
        \State \textbf{Input:} $ptr$, $IN$, $DA$, $K_h$, $K_w$, $O_h$, $O_w$
        \State \textbf{Output:} $O$
            % \For{$c = 0$ to $I_{c}$} 
            \For{each channel in $I_{c}$} 
            \IfThenOneLine {NPC \hspace{0.05em} \textrm{\textbf{then}}}% If ...2
            {$\text{Skip channel}$}% ...then...
            % % PNO Processing
            \State Initialize ptr\_sh = 0
            \State x = $ptr$[ptr\_sh] \Comment{First value of $ptr$ in each channel}
             \If {$x = 0$ \hspace{0.1em} \textrm{\textbf{then}}}
               % x 
               \For{x = $ptr$[ptr\_sh] to $ptr$[++ptr\_sh]}
                 \State convSpMv($K_h$,$K_w$,$O_h$, ,0,$x$,0,IN[x])
               \EndFor
               \For{x = $ptr$[ptr\_sh] to $ptr$[++ptr\_sh]}
                 \State convSpMv($K_h$,$K_w$,$O_h$, 0,$x$,  $O_w$-1,IN[x])
             \EndFor
               
             \EndIf % end NPO no skip
             
             % Rest of the pointers
             \State Skip all pointers with NPF 
             % Should be ceil(Kw/Sw)
              \For{$pType$ = $ptr$[ptr\_sh] to $K_w$} %\Comment{Rest of the Pointers}
              %\Comment{Rest of the Pointers}
                % If statement for skip
                \IfThenOneLine {NPF \hspace{0.05em} \textrm{\textbf{then}}}% If ...
                {$\text{Skip ptr}$}% ...then...
                % ...else...
                
                % Submat loop
                \State x = 0
                \For{$s = 0$ to $O_w$}

                % \Comment{Main Conv over Submats}
                  \IfThenOneLine {ptr[ptr\_sh+s+1]$<$0 \hspace{0.05em} \textrm{\textbf{then}}}   % If ... 
                   \State {$\text{ptr[ptr\_sh+s+1] = ptr[ptr\_sh+s]}$} 
                    % ...then...
                    % ...else...
                      
                  \For{$tx = x$ to $ptr$[ptr\_sh+s+1]}

                    \State convSpMv($K_h$,$K_w$,$O_h$,$pType$,$tx$, s,IN[tx])
                  \EndFor % x_ptr loop
                 
                 \State x = ptr[ptr\_sh+s+1], ptr\_sh++
                \EndFor % submats loop
                 
              \EndFor % end for the rest of the pointers
            \EndFor % end for channels
        \EndProcedure
    \end{algorithmic}
\end{algorithm}

% \vspace{-1em}

\subsection{CPS Encoding and Convolution Algorithms}
\label{sec:cps}

The goal of CPS is to produce further memory footprint savings over the CPO by exploiting the redundancy of neighbouring NZEs index positions. Over inception and residual CNNs, 81\% of the total NZEs appear in the overlap$K_w$ of the activation map that contribute the most to the output feature map. As a result, CPS executes the CPO to obtain $ptr$ and $DA$. For $IN$, CPS executes the CPO in all overlap regions except for the overlap$K_w$ region.

To obtain $IN_{CPS}$ from overlap$K_w$ region, CPS groups indices obtained in Section \ref{sec:alg_overview} in a number of sets of size 4, set4. Set4 is a set of 4 indices in the activation map's columns that belongs to overlap$K_w$ region. If $I_h$ is not a multiple of 4, then the last set4 will include the remaining indices and extra zeros to complete the set4. Each set4 can have $2^4 = 16$ index patterns depending on whether the indices inside of the set correspond to zero or non-zero data values. If the number of NZEs in set4 = 4 or 3, we encode the set4 as \{first NZE index of the set, pattern\}. Here, pattern could be 7 (0111), 11 (1011), 13 (1101), 14 (1110), or 15 (1111). The order of binary representations of each pattern is read from bottom to top in a given column. For example in Figure \ref{fig:org_feature_map_lowered_mat}, the first set4 in the second yellow column is \{3, 7, 11, 15\} which is encoded to \{7, 14\}. Here, 14 indicates 3 NZEs out of 4 with order 1110 read from bottom to top. If the number of NZEs in set4 is 2, we insert these two indices in $IN_{cps}$ after multiplying by -1. This multiplication is done to distinguish singletons from the case when set4 has 4 or 3 NZEs. If set4 has one NZE, we encode the singleton after multiplying by -1 (e.g., -23 as a singleton in Figure \ref{fig:org_feature_map_lowered_mat}). 
\footnote{Sets of size 4 are chosen to limit the number of pattern possibilities that consequently simplifies index reconstruction.}.

% sehs back
Algorithm \ref{alg:cpsEncodeAlg} executes the all NOP and overlapping types until overlap$K_w-1$ type as same as Algorithm \ref{alg:cpoEncodeAlg}, then CPS encoding executes the overlap$K_w$ region in lines 6-20. We determine the set4's pattern using the pattern variable in Line 10. After recognizing the number of NZEs (nz\_col) in the set4, the $IN_{CPS}$ will be encoded either to the first index of the recognized pattern if nz\_col $>$ 2 as in line 13 or singleton indices will be encoded from auxiliary vector (Aux) in Lines 15-18. 

Algorithm \ref{alg:cpsConvAlg} outlines the CPS convolution for overlap$K_w$ region that is done independently for each channel and kernel. In Figure \ref{fig:org_feature_map_lowered_mat}, we process overlap4 region, which has 1 NZE in partition A ($ptr$[8]). We read the corresponding index of this NZE i.e., -23. A negative index indicates a singleton that is reconstructed by turning it to positive. As a result, convSpMv is executed for the NZE 52 with index 23. In partition B, we have 4 NZEs ($ptr$[9] - $ptr$[8]) in $DA$ i.e., 13, 23, 33, and 53. We read the index 7 and pattern 14 from $IN_{cps}$, so there is 3 NZEs in this pattern. When patterns exist, we reconstruct the original index by adding the starting index with $K_w*nextOffset(pattern, g)$. For pattern 14 or 1110, we process 13 with index 7, 23 with index 7+$K_w$ = 11, 33 with index 7+$K_w$*2 = 15. The last NZE is 53 follows the same practice of the previous singleton (NZE = -23). More details are in Algorithm \ref{alg:cpsConvAlg}.

\begin{algorithm}
    \caption{CPS Encoding Algorithm}
    \label{alg:cpsEncodeAlg}
    \begin{algorithmic}[1]
        % \Function{CPS\_idx}{$IN, Aux, nz\_col, pattern, I_h, K_w, h, w$}
        %     \State \Return $IN$
        % \EndFunction
        \Procedure{cpsEncode}{}
            \State \textbf{Input:} $ FM$, $K_w$, $Pad\_Left$
            \State \textbf{Output:} $ptr$, $DA$, $IN$
            \State Initialize Aux[4], nz\_col,pattern = 0
            \State Insert Lines 10 to 26 From cpoEncode.
            \For{$w=K_w - 1$ to $I_w-K_w$} %\Comment{Last overlap$K_w$ region}
                \State Reset nz\_col, pattern to 0
                \For{$h$ = 0 to $I_h$}
                    \If{$I(h, w) \neq 0$ \hspace{0.1em} \textrm{\textbf{then}}} 
                    $DA$.insert($I(h,w)$),
                        % \State pattern += 1$\ll$($h$\%4), nz\_ptr++
                        \State pattern += $2^{(h\%4)}$, nz\_ptr++
                        \State Aux[++nz\_col]= -1 $\times$ ($w$ - idx + $h\times K_w$)
                        % \State CPS\_idx($IN, Aux, nz\_col,pattern, I_h, K_w, h, w$)
                        % \If{$(h+1)\%4$ = $0$}
                        % {
                            \If{nz\_col $>$ 2 \& ($h$\%3) = 0 \textrm{\textbf{then}}}
                                \State $IN$.insert($w$ - idx +$(h - (h\%4)) \times K_w$))
                                \State $IN$.insert(pattern)
                            \ElsIf{nz\_col = 2 \& ($h$\%3)$>$1 \hspace{0.1em} \textrm{\textbf{then}}}
                                \State $IN$.insert(Aux[1]), $IN$.insert(Aux[2])
                            \ElsIf{nz\_col = 1 \& ($h$\%3)$>$2 \hspace{0.1em} \textrm{\textbf{then}}}
                                \State $IN$.insert(Aux[1])
                            \EndIf
                            \IfThenOneLine{($h$\%3) = 0\hspace{0.1em} \textrm{\textbf{then}}} 
                            Reset nz\_col, pattern to 0
                            
                        % }
                        % \Else{} continue
                        % \EndIf
                    \EndIf
                \EndFor
                \State $ptr$[ptr\_sh] = nz\_ptr 
            \EndFor
            \State Insert Lines 29 to 36 From cpoEncode.
        \EndProcedure
    \end{algorithmic}
\end{algorithm}

{\setlength{\intextsep}{0pt}
\begin{algorithm}
    \caption{CPS Convolution Algorithm}
    \label{alg:cpsConvAlg}
    \begin{algorithmic}[1]
        \Procedure{cpsAlg}{}
        \State \textbf{Input:} $ptr$, $IN$, $DA$, $K_h$, $K_w$, $O_h$, $O_w$
        \State \textbf{Output:} $O$
            \For{each channel in $I_{c}$} 
             \IfThenOneLine {NPC/NPF \hspace{0.05em}    \textrm{\textbf{then}}}% If ...
             {$\text{Skip channel/ last ptr}$}% ...then...
                % Rest of the pointers
                %  \State Do all pointers via CPO except the last if no NPF
                % Submat loop
               
                % Submat loop
                \State x = 0
                \For{$s = 0$ to $O_w$}
                    \For{$tx = x$ to $ptr$[ptr\_sh+s+1]}. 
                    % \Comment{ptr is at overlap$K_w$}
                    
                    %     \IfThenOneLine {ptr[ptr\_sh+s + 1]$<$0 \hspace{0.05em} \textrm{\textbf{then}}}   % If ... 
                    % {$\text{ptr[ptr\_sh+s+1] = ptr[ptr\_sh+s]}$}
                    \IfThenOneLine {ptr[ptr\_sh+s+1]$<$0 \hspace{0.05em} \textrm{\textbf{then}}}   % If ... 
                    \State {$\text{ptr[ptr\_sh+s+1] = ptr[ptr\_sh+s]}$} 
                    
                    \State index = tmpIndex = IN[tx], pattern = -1
                    
                    \IfThenOneLine {index $<$ 0 \hspace{0.05em} \textrm{\textbf{then}}}% If ...
                    {$\text{index = IN[tx] = -1*IN[tx]}$}% ...then..
                    \State \textrm{\textbf{else}}
                    {$\text{pattern = IN[tx+1]}$}% ...then..
                    \State convSpMv($K_h$,$K_w$,$O_h$,$pType$,$tx$, s,IN[tx])
                    
                    \State nnzSet4 = getPatternSize(pattern, tmpIndex) 
                    
                    \For{$g = 1$ to $nnzSet4$}
                         \State index += $K_w$*nextOffset(pattern, g)
                         \State tx = tx + 1
                        
                        \State convSpMv($K_h$,$K_w$,$O_h$,$pType$, tx, s,index)
                    \EndFor % end for set size
                    
                 \EndFor % x_ptr loop
                    
                    \State x = ptr[ptr\_sh+s+1], ptr\_sh++
                \EndFor % submats loop
               
            \EndFor % end for channels
        \EndProcedure
    \end{algorithmic}
\end{algorithm}
}

\subsection{Hybrid Inference Over a Target CNN}
\label{sec:hybrid}

It is known from the field of linear algebra that the performance of SpMv depends on the input density \cite{park2016faster}. The lower the the density, the better SpMv's performance in time, and vice versa, which makes it challenging to integrate SpMv into all conv layers in real CNN inference. To alleviate this challenge and gain both time and memory savings of CPO/CPS in deep CNN inference, we carried out an experiment where we run 20 images from ImageNet validation set through the conv layers of Inception and Residual CNNs. For each conv layer, we record the average density across channels at each image. For ResNet-V2-152, the maximum variance of density over all layers is 0.006, which indicates that the density is approximately stationary from one image to another. This insight is consistent in other CNNs as shown in the SI and intuitive because weights are fixed and should produce stationary characteristics. 

Based on this insight, we embedded CPO/CPS into the CNN inference and synergized them using im2col. Suppose that the target CNN used for classification is known. We run our methods along with im2col on the CNN with $M$ images offline. For each layer, we record the total time in convolution of each of three convolution methods over the $M$ (M=5) images. For the given CNN, we have two modes of selection: favour time or favour space. Favour time modes selects the convolution method that yields the minimum total time between im2col and CPO, while favour space mode is between im2col and CPS. At inference time, the selected convolution method is used at this conv layer. This selection method results in a hybrid selection of CPO or CPS and im2col over conv layers.

%%%%%%%%%%%%%%%%%%%%%%%%%%%%%%%
\section{Experiments}
\label{sec:experimental_results}

This section demonstrates the effectiveness of our CPO and CPS written in single threaded C++ in terms of time and memory savings. We report the results relative to im2col used in Caffe framework since im2col is the most popular algorithm \cite{hadjis2015caffe}. In addition, we wrote the implementation of CSCC in \cite{fan2019cscc}. Also, same conv layer settings are used to compare CPO/CPS with either COO-based algorithm or MEC implemented in parallel in \cite{shi2017speeding} and \cite{cho2017mec}, respectively. Due to limited resources, 100 randomly sampled images from the ImageNet dataset were used as input stimuli \footnote{During code development, randomized feature maps were generated to verify our results as well.} \cite{krizhevsky2012imagenet}.

We validate our algorithms on 6 CNNs: IV1, IV3, Inception V4 (IV4), ResNet-V2-50,  ResNet-V2-101, and ResNet-V2-152 \cite{szegedy2015going, szegedy2016inceptionv3, szegedy2017inceptionv4, he2016identityresnet}. From these CNNs, we show two sets of results: 1) per-layer results, 2) end-to-end inference results. For the first set, a sample of 13 conv layers and various characteristics is shown out of 252 conv layers that can be considered by CPO/CPS. We encourage the reader to view the rest of the conv layers results in the SI. To accurately measure the time savings percentage, we calculate the average wall clock time of all algorithms under test after 100 iterations. For the second set, we choose another 5 randomly sampled images for the hybrid convolution selection method. Then, we compute the default total time taken by all types of layers in each CNN under test and compare this time with the total time taken by the selected methods using our proposed hybrid approach. All conv layers not using CPO/CPS are executed using their default implementations. Experiments are carried out on a 16GB memory Quad-Core Intel i7-4790 (3.60 GHz)

\begin{figure*}[!ht]
  \begin{minipage}{\textwidth}
  % without a b c 
  \captionsetup[subfigure]{labelformat=empty}
  \centering
    \subfloat[Speed/Memory IV1 \label{fig:IV4}][]{\includegraphics[height=0.55\textwidth, width=1\textwidth]{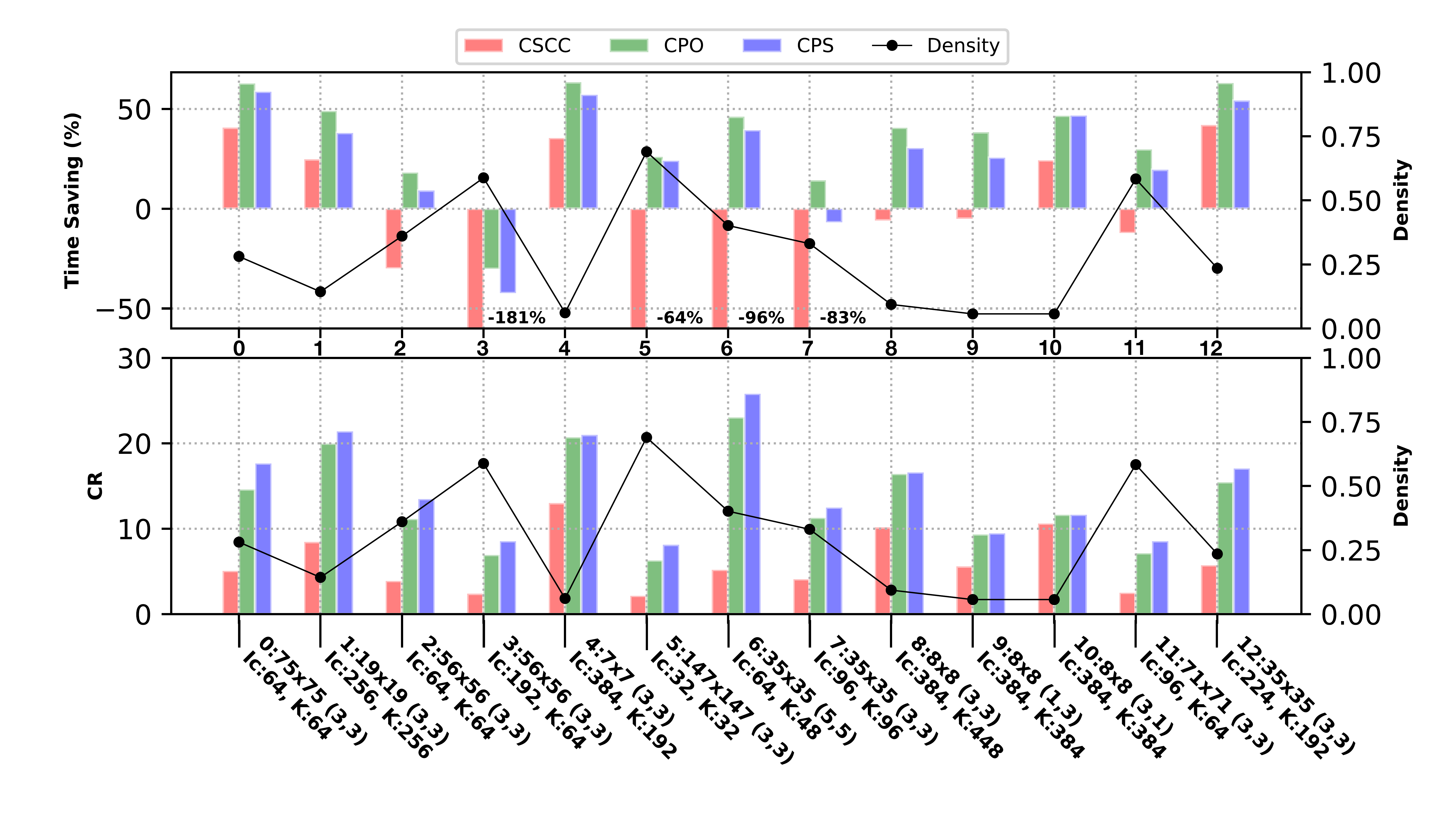}}
  \end{minipage}
  \vspace{-3em}
  \caption{Per-layer Performance Results of CPO, CPS, and CSCC. layers are in this format: layerID: $I_h\times I_w$, ($K_h\times K_w$)}
  \label{fig:speed_memory_cpo_cps_test}
\end{figure*}

\begin{table*}[ht]
\caption{Inference time and memory savings using the proposed hybrid approach based on CPO/CPS. \textit{Part:} Savings in non-pointwise layers with $s_h=s_w=1$. \textit{E2E:} Overall inference end-to-end savings.}
\label{tab:hybrid_selection_results}
\vspace{-1em}
\begin{center}
\setlength{\intextsep}{-50ex}
\resizebox{1.80\columnwidth}{!}{%
\begin{tabularx}{\linewidth}{? Y ? c | c ? c | c ? c | c ? c | c ? c | c ? c | c ? }
    \specialrule{.1em}{.05em}{.05em} 
     \multirow{2}{*}{\diagbox[width=23mm, linewidth=1.5pt ]{\textbf{Results}}{\textbf{CNN}}} &
    \multicolumn{2}{c?}{\textbf{Resnet-V2-50}} &  \multicolumn{2}{c?}{\textbf{Resnet-V2-101}} &  \multicolumn{2}{c?}{\textbf{Resnet-V2-152}} &  \multicolumn{2}{c?}{\textbf{IV1}} & \multicolumn{2}{c?}{\textbf{IV3}} & \multicolumn{2}{c?}{\textbf{IV4}}
    \\
     \clineB{2-13}{3.5}
    & Part & E2E & Part & E2E & Part & E2E & Part & E2E & Part & E2E & Part & E2E \\ 
    \specialrule{.1em}{.05em}{.05em}  
   \hline
    \textbf{\textbf{T$_{cpo}$(\%)}} & 18 & 6.9 & 21.4 & 7.1 & 12.1 & 1.8 & -15.9 & -8 & -1 & -0.6 & -0.5 & -0.3 \\ 
    \specialrule{.1em}{.05em}{.05em} 
    \hline
    \textbf{\textbf{T$_{cps}$(\%)}} & 2.3 & 1 & 6.5 & 2.1 & -4.5 & -0.7 & -35.1 & -17.6 & -11.8 & -7.1 & -11.1 & -6.1 \\
    \specialrule{.1em}{.05em}{.05em}  \hline
    % \textbf{\textbf{T$_{opTime}$(\%)}} & 20.8 & 8 & 23.5&7.7 & 18.2 & 2.8 & 9.4 & 4.7 & 14.7& 8.9 & 15.6 & 8.5 \\ 
    % \specialrule{.1em}{.05em}{.05em}  \hline
    \textbf{\textbf{T$_{hybCPO}$(\%)}} & 20.2 & 7.8 & 22.9 & 7.5 & 17.9 & 2.7 & 9.1 & 4.6 & 14.6 & 8.8 & 14.7 & 8 \\
    \specialrule{.1em}{.05em}{.05em}  \hline
    \textbf{\textbf{T$_{hybCPS}$(\%)}} & 10.2 & 4 & 12.2 & 4 & 9.8 & 1.5 & 6.6 & 3.3 & 9.4 & 5.7 & 8.6 & 4.7 \\
    \specialrule{.1em}{.05em}{.05em} 
    \specialrule{.1em}{.05em}{.05em}  \hline
    \textbf{\textbf{CR$_{cpo}$}} & 12.2x & 9.6x & 13.3x & 11.9x & 13.3x & 12.4x & 9.4x & 9x & 9.7x & 8.9x & 9.7x & 9x \\ 
    \specialrule{.1em}{.05em}{.05em}  \hline
    \textbf{\textbf{CR$_{cps}$}} & 13.5x & 10.5x & 14.6x & 13x & 14.3x & 13.2x & 10.3x & 7.8x & 10.3x & 9.4x & 10.2x & 9.47x \\
    \specialrule{.1em}{.05em}{.05em}  \hline
    % \textbf{\textbf{CR$_{opTime}$}} & 10x & 7.9x & 10.8x & 9.7x & 9.6x & 8.9x & 3.6x & 3.5x & 5.4x & 5x & 6x & 5.62x \\ \specialrule{.1em}{.05em}{.05em}  \hline
    \textbf{\textbf{CR$_{hybCPO}$}} & 9.7x & 7.7x & 11.6x & 10.3x & 10.3x & 9.6x & 3.5x & 3.4x & 6x & 5.5x & 6.3x & 5.9x \\
    \specialrule{.1em}{.05em}{.05em}  \hline
    \textbf{\textbf{CR$_{hybCPS}$}} & 10.6x & 8.4x & 9.8x & 8.8x & 8x & 7.4x & 2.9x & 2.8x & 4.8x & 4.4x & 5.97x & 5.6x \\
    \specialrule{.1em}{.05em}{.05em}  
\end{tabularx}
}
\end{center}
\vspace{-2em}
\end{table*}

\textbf{CPO/CPS per-layer performance results relative to im2col:} Figure \ref{fig:speed_memory_cpo_cps_test} shows the performance comparisons in terms of time and memory of CPO and CPS. As shown in this figure, the average CR for CPO and CPS across these layers is 13.3x and 14.7x, respectively. Also, CPO/CPS achieves high per-layer CR savings up to 25.8x. For example, Layer 6 (density of 0.4, 35x35 activation map, and 3x3 kernel) has CR of 23x and 25.8x for CPO/CPS, respectively. For time savings, CPO/CPS may not do well in some layers with high density due to the SpMv memory challenges reported in the literature \cite{szegedy2015going, park2016faster}. For instance, Layer 3 has a density of $\approx$0.6 and our time losses equal to 30\% for CPO and 42\% for CPS, but our CR gains are 6.9x for CPO and 8.5x for CPS. However, the average time savings of CPO and CPS in these layers is 36\% and 27\%, respectively. In addition, CPO/CPS achieves on average time savings up to 63.4\%. For example, Layer 4 has a density of 0.06 and time savings are 63.4\%, 57\% for CPO, CPS, respectively. Layer 11 is another example, where the average time saving is 30\%, 20\% for CPO, CPS, respectively. These savings results are obtained, while preserving the classification accuracy of CNNs under test.

\textbf{Comparison between CPO/CPS and CSCC:}
Similar to what's shown in \cite{fan2019cscc}, CSCC's time saving and CR in layer 4 are better than im2col (35.4\% and 13x, respectively). However, supported by Figure \ref{fig:speed_memory_cpo_cps_test}, CPO/CPS time and memory results in all layers are better than CSCC.

\textbf{Comparison between CPO/CPS and COO:} Inception4e.3 (density=0.1) and Inception5b.3 (density=0.05) in \cite{shi2017speeding} are used to compare our CPO/CPS with COO. In these layers, the average density over 100 images was more than the density reported in \cite{shi2017speeding}. Thus, we choose images in our input stimuli with similar density to \cite{shi2017speeding}. For Inception4e.3 and an image with density 0.16, CPO, CPS, and \cite{shi2017speeding} achieve time savings of 34.3\%, 17.5\%, 37.1\%, respectively, while CR is 16.8x, 17.6x, and 17.2x, respectively. For Inception5b.3 and an image with density 0.05, CPO, CPS, and \cite{shi2017speeding} achieve time savings of 74.5\%, 68.6\%, 78.89\%, respectively, while CR is 31.81x, 32.81x, 27.2x, respectively. For Inception5b.3 and an image with density 0.013, we can achieve time saving of 90.4\%, 88.2\% and CR of 62.4x, 62.6x for CPO and CPS respectively. From these results, CPO/CPS can outperform the COO in terms of CR with a minor drop in time savings. It is worth noting that we apply our proposed CPO/CPS to real CNN inference and examine CPO/CPS on layers at different densities rather than only at 0.05 and 0.1.

\textbf{Comparison between CPO/CPS and parallel MEC:} Per-layer comparisons are conducted using 4 conv layers already existing in the CNNs under test: cv9, cv10, cv11, cv12 in \cite{cho2017mec}. MEC performed better in term of time savings in cv9 and cv10, yet CPO/CPS outperformed MEC in terms of CR. We conjecture that these layers do not have enough sparsity degree with respect to their activation map size. For cv11 with density of 0.12, MEC's time saving is 26.36\% and CR is 2.57x, CPO's time saving is 54.16\% and CR is 22.87x, and CPS's time saving is 44.87\% and CR is 23.66x. For cv12 with density of 0.13, MEC's time saving and CR are 36.97\% and 2.16x, respectively. CPO's time saving for this layer is 37.34\%, and CR is 16.16x, while CPS's time saving is 26.54\%, and CR is 16.48x. From these results, 
CPO/CPS can outperform the parallel MEC's performance in some layers.

\textbf{Practical implications of CPO/CPS on CNN Inference:} 
In the first part of Section \ref{sec:experimental_results}, CPO/CPS improved the performance of im2col in some layers, but with some drops in terms of time savings. Thus, we apply the proposed hybrid convolution selection method in Section \ref{sec:hybrid}.

Table \ref{tab:hybrid_selection_results} provides the time savings as well average CR of CPO, CPS, hybrid with CPO and im2col (hybCPO), and hybrid with CPS and im2col (hybCPS). For each CNN, we report time savings in the portion the CPO/CPS is operating (Part) and end-to-end time (E2E) savings. For example, CPO on its own can improve the time of ResNet-V2-101 by 21.4\% over the layers CPO is used and by 7.1\% in end-to-end. In the same case, the CR is 13.31x/11.9x, respectively. Also, we can observe that there are insignificant time losess in some of the CNNs under test when using only CPO/CPS, but there are CR gains that may be useful for deploying deep CNNs on memory-constrained devices with limited bandwidth. Eliminating these losses are done via the hybrid selection approach. 

In hybCPO and hybCPS, we calculate the percentage of layers that CPO was selected among the layers that are not pointwise convolution. In hybCPO, CPO was selected with a percentage of 56\% on average across CNNs, while CPS was selected with a percentage of 35.5\% on average in hybCPS. Hybrid CPO and CPS save inference time with a percentage 
up to $\approx$9\% and up to 6\%, respectively, while the average CR across layers of hybCPO and CPS are up to 10.3x and 9x, respectively. CR of hybCPS is less than CPO because it is selected less than CPO, yet hybCPS yields that maximum per-layer CR gains when CPS is selected. These results demonstrate the potential of CPO/CPS in CNN inference.

%%%%%%%%%%%%%%%%%%%%%%%%%%%%%%

\section{Conclusion and Future Work}
\label{sec:conlusion_future_work}

This paper proposed CPO and CPS convolution that simultaneously decrease the memory footprint and increase the inference speed while preserving the accuracy. We demonstrated the effectiveness of CPO and CPS with per-layer time savings up to 63\% and CR up to 26x with respect to im2col. To the best of our knowledge, this is the first paper that showed both the time and space savings of its algorithms in CNN inference. Using the proposed hybrid convolution selection method, end-to-end inference time savings go up to 9\% and CR savings up to 10x. Future extensions could be parallel sparse methods for training/inference on CPUs/GPUs or light-weight CNNs. 

{\small
\bibliographystyle{ieee_fullname}
\bibliography{egbib}
}

% Add SI
\onecolumn

\input{supplement}

\end{document}

%% file: supplement.tex
% Example definitions.
% --------------------
\def\x{{\mathbf x}}
\def\L{{\cal L}}

\begin{center}
\textbf{\Large Supplementary Information}
\end{center}

\appendix

\section{Space Complexity Derivations of CPO/CPS}
\label{sec:derivations}

% This subsection derives the space complexity of our CPO/CPS encoding representations for the already considered non-pointwise convolutional layers with $s_h=s_w=1$. Then, we compare CPO/CPS space complexity with im2col, MEC, and CSCC.

This subsection derives the space complexity of our CPO/CPS encoding representations. Then, we compare CPO/CPS space complexity with im2col, MEC, and CSCC. For each of im2col in subsection 1.1, MEC in subsection 1.2, and CSCC in subsection 1.3, we derive density bounds where the space complexity of our CPO/CPS is better.

% for which we calculate the denbounds 

The size of Overlapping Pointer (OP) is defined as follows:

\begin{equation}
OP\_size =
     \begin{cases}
      (1 + O_{w}) & \text{If\:}\, K_w = 1, S_w = 1 \\
      (\lceil{\frac{K_{w}}{s_{w}}}\rceil - M)(1 + O_{w}) & \quad\text{Otherwise} \\ 
      %(\frac{K_{w}}{s_{w}} - M)(1 + O_{w}) &\quad\text{otherwise}
     \end{cases}
     \label{eq:kw_sw_frac}
\end{equation}

\noindent where $M = max(1, pad\_left)$. The size of Non Overlapping Pointer (NOP) is 3 if the padding is VALID. 
%A 0 element is first added to indicate the NOP type. Also, the first and last column that appear in the non-overlapping region might contain NZEs for which we need 2 other elements. The size of the non-overlapping pointer $NOP$ is 0 if it is SAME padding. 
First element indicates NOP type while second and third elements represent the number of NZEs of the first and  the last columns, respectively. When the padding is SAME, the size of NOP is zero. The total size of $ptr$ is defined as follows:
% \vspace{-10pt}

\begin{flalign}
ptr\_size  = \sum_{n=1}^{I_{n}} \sum_{c=1}^{I_{c}} 
  \max(1, \mathbf{1}_{\rho_{n,c}>0} ( NOP\_{size} + OP\_{size})) -count_{NPF}*(O_w-1) 
\label{eq:ptrSize}
\end{flalign}
where $\rho_{n,c}$ is the density of the cth channel of activation map for an image $n$ and $count_{NPF}$ is the total count of NPF skip flags depending on the zero and non-zero elements distribution. The size of $DA_{CPO}$, $DA_{CPS}$, and $IN_{CPO}$ are as follows:
\begin{equation}
DA\_size_{CPO} = DA\_size_{CPS} =  IN\_size_{CPO} = I_{h}I_{w}\sum_{n=1}^{I_{n}} \sum_{c=1}^{I_{c}} \rho_{n,c}
\label{eq:data_index_size_cpo}
\end{equation}
However the size of the $IN_{CPS}$ is as follows:

\begin{flalign}
IN\_size_{CPS} = \sum_{n=1}^{I_{n}} \sum_{c=1}^{I_{c}} (C'_{1_{n,c}}+2C'_{2_{n,c}}+3C'_{3_{n,c}}+ 4C'_{4_{n,c}} + C''_{1_{n,c}}+2C''_{2_{n,c}}+2C''_{3_{n,c}}+ 2C''_{4_{n,c}})
\label{eq:index_size_cps}
\end{flalign}
where $C''_{k_{n,c}}$ indicates the number of set4 which has $k$ NZEs inside of the set4 for the cth channel of activation map of an image $n$ in overlap$K_{w}$ region.
%the NZE total count in overlap$K_w$ region of activation maps,
$C'_{k_{n,c}}$ represents the number of set4 which has $k$ NZEs inside of the set4 for the cth channel of activation map of an image $n$ in overlap2 type until overlap$(K_{w}-1)$ as well as NOP regions. 
%the count in the rest of overlap regions.

Finally, the total size of CPO/CPS structure is the summation of their corresponding $ptr$, $DA$, and $IN$ sizes.

\subsection{Space Complexity Difference between CPO and im2col:}

We can define the space complexity of im2col  \cite{chellapilla2006high_im2col}, $S_{im2col}$ as follows:  
  \begin{flalign}
 S_{im2col} = I_n O_w O_h I_c K_w K_h = I_n (\lceil \frac{I_w - K_w}{s_w} \rceil +1) (\lceil \frac{I_h - K_h}{s_h} \rceil +1) I_c K_w K_h. \label{eq1-5}
 \end{flalign}
To compute the difference in terms of memory units between im2col, $S_{im2col}$, and CPO in the worse case, $S_{CPO}$ ($M = 1$, VALID padding, no NPC, no NPF, and $K_{w} > s_{w}$), we have:
\begin{flalign}
 S_{im2col} - S_{CPO} =  I_n O_w O_h I_c K_w K_h -I_{n}I_{c} \left( 3 + (\lceil{\frac{K_{w}}{s_{w}}}\rceil - 1)(1 + O_{w})\right) - 2I_{h}I_{w}\sum_{n=1}^{I_{n}} \sum_{c=1}^{I_{c}}\rho_{n,c}
\end{flalign}
To ensure $S_{im2col} - S_{CPO} \geq 0$, we have to satisfy the following equation:
\begin{flalign}
2 I_h I_w \sum_{n=1}^{I_{n}} \sum_{c=1}^{I_{c}}\rho_{n,c} & \leq I_n O_w O_h I_c K_w K_h + I_{n}I_{c} (O_{w}+1) - \left( 3+\lceil{\frac{K_{w}}{s_{w}}}\rceil (1+O_{w}) \right) I_n I_c \nonumber \\
\sum_{n=1}^{I_{n}} \sum_{c=1}^{I_{c}}\rho_{n,c} & \leq \frac{I_{n}I_{c}}{2 I_h I_w} \left ( O_w (K_w K_h O_h+1-\lceil{\frac{K_{w}}{s_{w}}}\rceil) - 2 - \lceil{\frac{K_{w}}{s_{w}}}\rceil\right )
\end{flalign}
Then, we can deduce the following:
\begin{flalign}
\forall_{S_{CPO}, S_{im2col}}(S_{im2col} - S_{CPO}) \geq 0 \Leftrightarrow  \exists_{\rho} P(\rho) 
\end{flalign}
% \newpage
where $\rho = \sum_{n=1}^{I_{n}} \sum_{c=1}^{I_{c}}\rho_{n,c}$ and  

\begin{flalign}
P(\rho)= \{\rho | 0 \leq \rho \leq min(1, max(0, \frac{I_{n}I_{c}}{2 I_h I_w} \left ( O_w (K_w K_h O_h+1-\lceil{\frac{K_{w}}{s_{w}}}\rceil) - 2 - \lceil{\frac{K_{w}}{s_{w}}}\rceil\right ))\}.
\end{flalign}

%Since the upper density bound is greater than 0.5 in typical CNNs, CPO works better than im2col in terms of space complexity.

\subsection{Space Complexity Difference between CPO and MEC:}

We can define the space complexity of \textbf{Memory Efficient Convolution (MEC)}, $S_{MEC}$, in \cite{cho2017mec} as follows:  
 \begin{flalign}
 S_{MEC} = I_n O_w K_w I_h I_c 
 &= I_n (\lceil \frac{I_w - K_w}{s_w} \rceil +1) K_w I_h I_c. \label{eq1-3}
 \end{flalign}
Now, we compute the difference in terms of memory units between MEC, $S_{MEC}$, and CPO in the worse case, $S_{CPO}$ ($M = 1$, VALID padding, no NPC, no NPF, and $K_{w} > s_{w}$):
\begin{flalign}
 S_{MEC} - S_{CPO} = I_n I_c O_w K_w I_h  -I_{n}I_{c} \left( 3 + (\lceil{\frac{K_{w}}{s_{w}}}\rceil - 1)(1 + O_{w})\right) - 2I_{h}I_{w}\sum_{n=1}^{I_{n}} \sum_{c=1}^{I_{c}}\rho_{n,c}
\end{flalign}
To ensure $S_{MEC} - S_{CPO} \geq 0$, we have to satisfy the following equation:
\begin{flalign}
2 I_h I_w \sum_{n=1}^{I_{n}} \sum_{c=1}^{I_{c}}\rho_{n,c} & \leq I_n I_c O_w K_w I_h  + I_{n}I_{c} (O_{w}+1) - \left( 3+\lceil{\frac{K_{w}}{s_{w}}}\rceil (1+O_{w}) \right) I_n I_c \nonumber \\
\sum_{n=1}^{I_{n}} \sum_{c=1}^{I_{c}}\rho_{n,c} & \leq \frac{I_{n}I_{c}}{2 I_h I_w} \left ( O_w (K_w I_h+1-\lceil{\frac{K_{w}}{s_{w}}}\rceil) - 2 - \lceil{\frac{K_{w}}{s_{w}}}\rceil\right )
\end{flalign}
Then, we can deduce the following:
\begin{flalign}
\forall_{S_{CPO}, S_{MEC}}(S_{MEC} - S_{CPO}) \geq 0 \Leftrightarrow  \exists_{\rho} P(\rho) 
\end{flalign}
% \newpage
where $\rho = \sum_{n=1}^{I_{n}} \sum_{c=1}^{I_{c}}\rho_{n,c}$ and  

\begin{flalign}
P(\rho)= \{\rho | 0 \leq \rho \leq min(1, max(0, \frac{I_{n}I_{c}}{2 I_h I_w} \left ( O_w (K_w I_h+1-\lceil{\frac{K_{w}}{s_{w}}}\rceil) - 2 - \lceil{\frac{K_{w}}{s_{w}}}\rceil\right ))\}.
\end{flalign}

\subsection{\textbf{Space Complexity Difference between CPO and CSCC:}}

To define the space complexity of \textbf{Convolution Split Compression Calculation (CSCC)}, $S_{CSCC}$, in \cite{fan2019cscc} let's define $\hat{\rho}_i:$

 \begin{flalign}
 \hat{\rho_i} = \frac{\sum_{i=1}^{K_w I_h I_c}\sum_{j=1}^{O_w}\mathbf{1}_{LM_{i, j} \neq 0}}{O_w K_w I_h I_c}
 \end{flalign}
 
 \noindent where, $\hat{\rho}_i$ $\in [0, 1]$ is the density of the lowered matrix $(LM)_i$. 
 
 Based on $\hat{\rho_i}$, we define the density $\hat{\rho}$ for the batch of the lowered matrices for input feature maps as follows:
 \begin{flalign}
 \hat{\rho} = \sum_{i=1}^{I_n} \hat{\rho}_{i}
 \end{flalign}
 
 \begin{flalign}
 S_{CSCC} & = I_{n} I_{c} (O_w + 1) + 2 \: O_w K_w I_h I_c \hat{\rho} \nonumber \\
 & = I_{n} I_{c} (\lceil \frac{I_w - K_w}{s_w} \rceil +2) + 2 \: (\lceil \frac{I_w - K_w}{s_w} \rceil +1) K_w I_h I_c \hat{\rho} \label{eq1-4}
 \end{flalign}
 
%  \textbf{\rom{1}) Difference between CPO and CSCC: }
% For comparison, let us form their different:
For comparison, let us calculate the difference between CSCC, $S_{CSCC}$, and CPO, $S_{CPO}$ in the worst case ($M = 1$, VALID padding, no NPC, no NPF, and $K_{w} > s_{w}$):
\begin{flalign}
 S_{CSCC} - S_{CPO}   = I_n I_c (O_w + 1) + 2 \:O_w I_h K_w I_c \hat{\rho}
  - I_{n}I_{c} ( 3 + (\lceil{\frac{K_{w}}{s_{w}}}\rceil - 1)(1 + O_{w}))  - 2I_{h}I_{w}\sum_{n=1}^{I_{n}} \sum_{c=1}^{I_{c}}\rho_{n,c} 
\end{flalign}
This difference will be non-negative if and only if,
\begin{flalign}
2I_{h}(O_{w}K_{w}I_{c}\hat{\rho}-I_{w}\sum_{n=1}^{I_{n}} \sum_{c=1}^{I_{c}}\rho_{n,c}) &\geq -I_{n}I_{c}(O_{w}+1)(1-(\lceil{\frac{K_{w}}{s_{w}}}\rceil - 1))+3I_{n}I_{c} \nonumber\\ 
O_{w}K_{w}I_{c}\hat{\rho}-I_{w}\sum_{n=1}^{I_{n}} \sum_{c=1}^{I_{c}}\rho_{n,c} &\geq \frac{1}{2I_{h}} (-I_{n}I_{c}(O_{w}+1)(2-\lceil{\frac{K_{w}}{s_{w}}}\rceil ))+3I_{n}I_{c})  \nonumber \\ -I_{w}\sum_{n=1}^{I_{n}} \sum_{c=1}^{I_{c}}\rho_{n,c} &\geq -O_{w}K_{w}I_{c}\hat{\rho} - \frac{1}{2I_{h}} (I_{n}I_{c}(O_{w}+1)(2-\lceil{\frac{K_{w}}{s_{w}}}\rceil)-3I_{n}I_{c}) \nonumber \\
\sum_{n=1}^{I_{n}} \sum_{c=1}^{I_{c}}\rho_{n,c}  & \leq \frac{1}{I_{w}} \Big(O_{w}K_{w}I_{c}\hat{\rho} + \frac{1}{2I_{h}} (I_{n}I_{c}(O_{w}+1)(2-\lceil{\frac{K_{w}}{s_{w}}}\rceil)-3I_{n}I_{c})\Big)
\end{flalign}

Then, we can deduce the following:

% Hossam: needs modification
\begin{flalign}
\forall_{S_{CPO}, S_{CSCC}} (S_{CSCC} - S_{CPO}) \geq 0 \Leftrightarrow  \exists_{\rho} P(\rho) \land \exists_{\hat{\rho}} Q(\hat{\rho})\\ \nonumber
\end{flalign}
where $\rho = \sum_{n=1}^{I_{n}} \sum_{c=1}^{I_{c}}\rho_{n,c}$, 
\begin{flalign}
P(\rho)= \{\rho | 0 \leq \rho \leq  min(1, max(0, \frac{1}{I_{w}} \Big(O_{w}K_{w}I_{c}\hat{\rho} 
+ \frac{1}{2I_{h}} (I_{n}I_{c}(O_{w}+1)(2-\lceil{\frac{K_{w}}{s_{w}}}\rceil)-3I_{n}I_{c})\Big)
\end{flalign}
, and
\begin{flalign}
Q(\hat{\rho}) = \{\hat{\rho} | 
\hat{\rho} \geq \frac{-I_{n}I_{c}(O_{w}+1)(2-\lceil{\frac{K_{w}}{s_{w}}}\rceil)+3I_{n}I_{c})}{2 \: O_w I_c I_h K_w} \}
\end{flalign}

For CPS, note the following: 
\begin{flalign}
\label{eq:cps_cpo_relation}
I_h I_w \sum_{n=1}^{I_{n}} \sum_{c=1}^{I_{c}} \rho_{n,c} & = \sum_{n=1}^{I_{n}} \sum_{c=1}^{I_{c}}(C'_{1_{n,c}}+2C'_{2_{n,c}}+3C'_{3_{n,c}} 
+ 4C'_{4_{n,c}} + C''_{1_{n,c}}+2C''_{2_{n,c}}+3C''_{3_{n,c}} + 4C''_{4_{n,c}} )
\nonumber \\
&  \geq 
\sum_{n=1}^{I_{n}} \sum_{c=1}^{I_{c}} (C'_{1_{n,c}}+2C'_{2_{n,c}}+3C'_{3_{n,c}} + 4C'_{4_{n,c}} 
+ C''_{1_{n,c}}+2C''_{2_{n,c}}+2C''_{3_{n,c}}+ 2C''_{4_{n,c}} )
\end{flalign}

\noindent which shows that the space savings of CPS relative to im2col can be greater than or equal to the space savings of CPO relative to im2col. 

As seen by Equation \ref{eq:cps_cpo_relation}, the space savings of CPS depend on the distribution of the zero and non-zero elements in the activation map. In the previous subsubsections, we showed that the space savings of CPO could be greater than that of other methods in the literature depending on the density of the activation map, and other spatial properties of the activation map as well as the kernel. This may show indirectly that the space savings of CPS are also greater than other methods in some convolutional layers depending on their characteristics. This can also be verified by the experimental results presented in the paper.

%%%%%%%%%%%%%%

%\pagebreak

\section{Detailed Analysis of our Experimental Results}
\label{sec:detailed}

In the paper, we have already shown a representative sample of per-layer results as well as overall inference results using CPO/CPS in all CNNs under the test. To achieve transparency, this subsection shows all per-layer results in all CNNs under the test of CPO/CPS. Please note that all these layers were already investigated in the paper via the overall inference results. Results in this subsection are reported with respect to im2col for non-point wise convolution layers with $s_h=s_w=1$.

Tables \ref{tab:long4}, \ref{tab:long5}, \ref{tab:long6}, \ref{tab:long1}, \ref{tab:long2}, and \ref{tab:long3} show the characteristics of the already considered non-point wise convolution layers with $s_h=s_w=1$. A row in each table shows the layer ID, the corresponding activation map height ($I_h$), the corresponding activation map width ($I_w$), the corresponding output height ($O_h$), the corresponding output width ($O_w$), the corresponding kernel height ($K_h$), the corresponding kernel width ($K_w$), the corresponding vertical stride ($s_h$), the corresponding horizontal stride ($s_w$), the corresponding number of channels ($I_c$), and the corresponding number of kernels ($K$).

\begin{figure*}[]

  \begin{minipage}{\textwidth}
  % without a b c 
  \captionsetup[subfigure]{labelformat=empty}
  \centering
    \subfloat[Density ResNet-50-V2
    \label{fig:IV1}][]{\includegraphics[width=1\textwidth,height=0.4\textwidth]{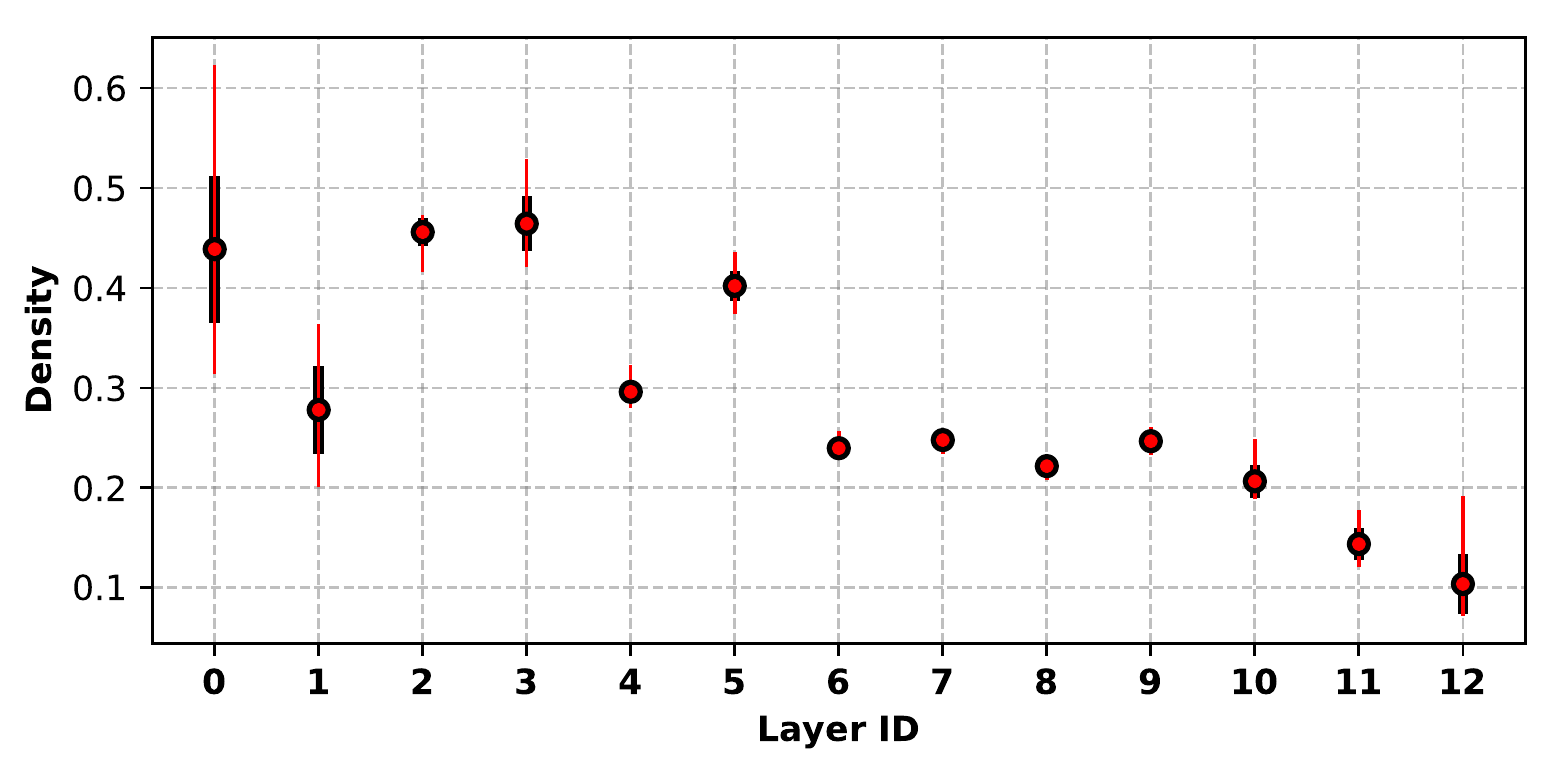}} 
    \\
    \subfloat[Speed IV1 \label{fig:IV1}][]{\includegraphics[width=1\textwidth,height=0.4\textwidth]{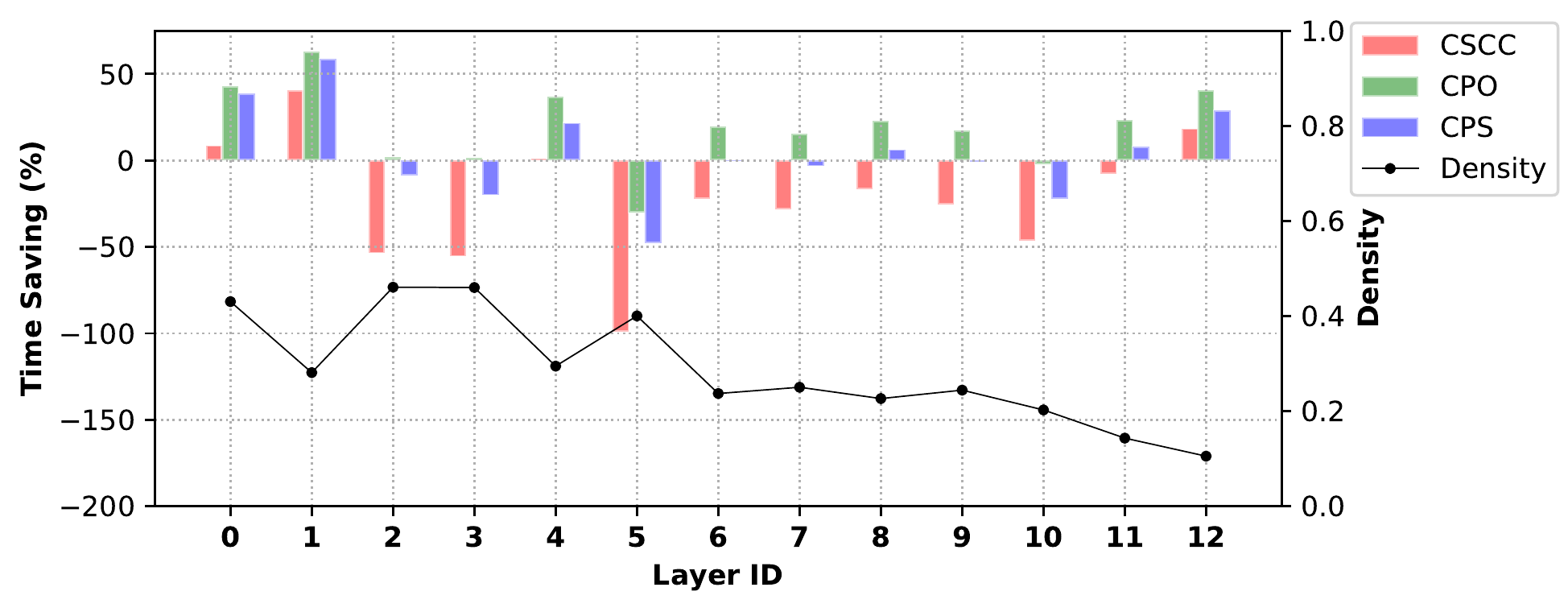}} \quad
    \subfloat[Memory IV1 \label{fig:IV1}][]{\includegraphics[width=1\textwidth,height=0.4\textwidth]{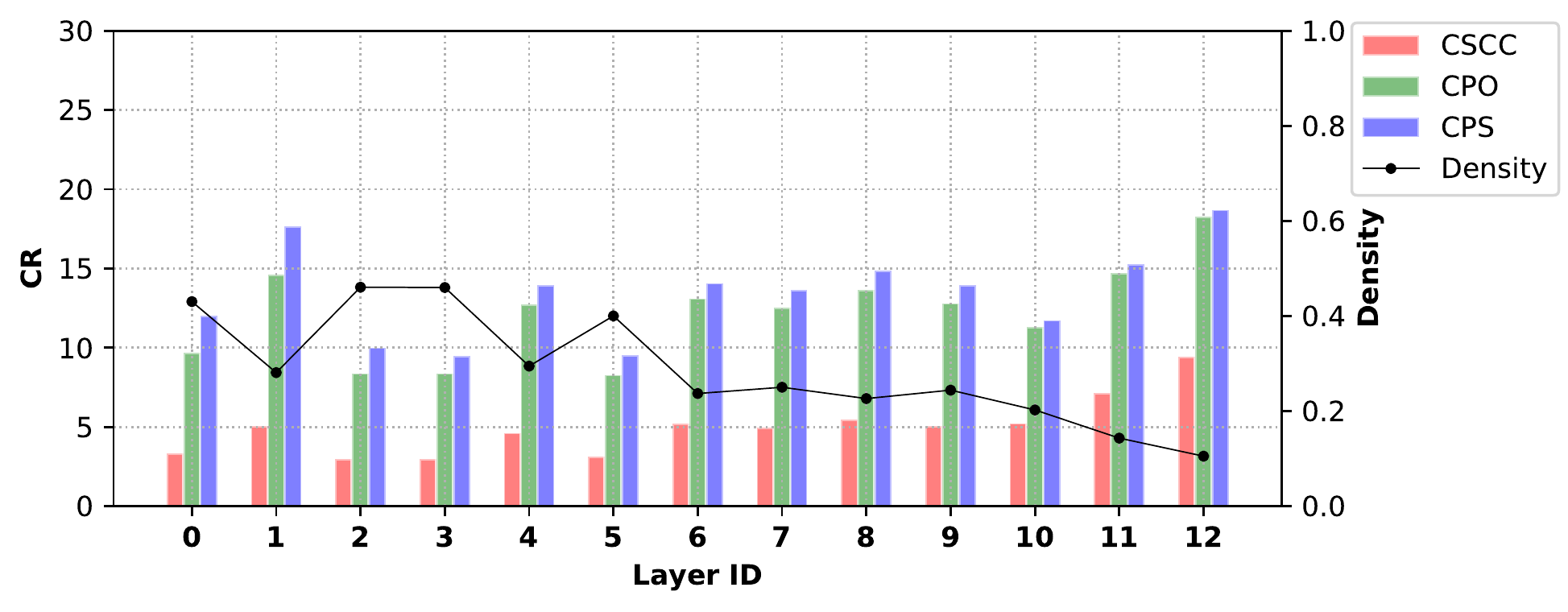}}
  \end{minipage}
%   \vspace{-3em}
  \caption{Top to bottom: Per-layer Stationary Density in \textit{ResNet-V2-50} for 20 Images, Per-layer Time Saving Results of CPO, CPS, and CSCC  in \textit{ResNet-V2-50} for 100 Images, Per-layer CR Results of CPO, CPS, and CSCC  in \textit{ResNet-V2-50} for 100 Images}
  \label{fig:density_speed_space_ResNet_50}
\end{figure*}

\begin{sidewaysfigure}[]
  \begin{minipage}{\textwidth}
  % without a b c 
  \captionsetup[subfigure]{labelformat=empty}
  \centering
    \subfloat[Density IV3 \label{fig:IV1}][]{\includegraphics[width=1.1\textheight]{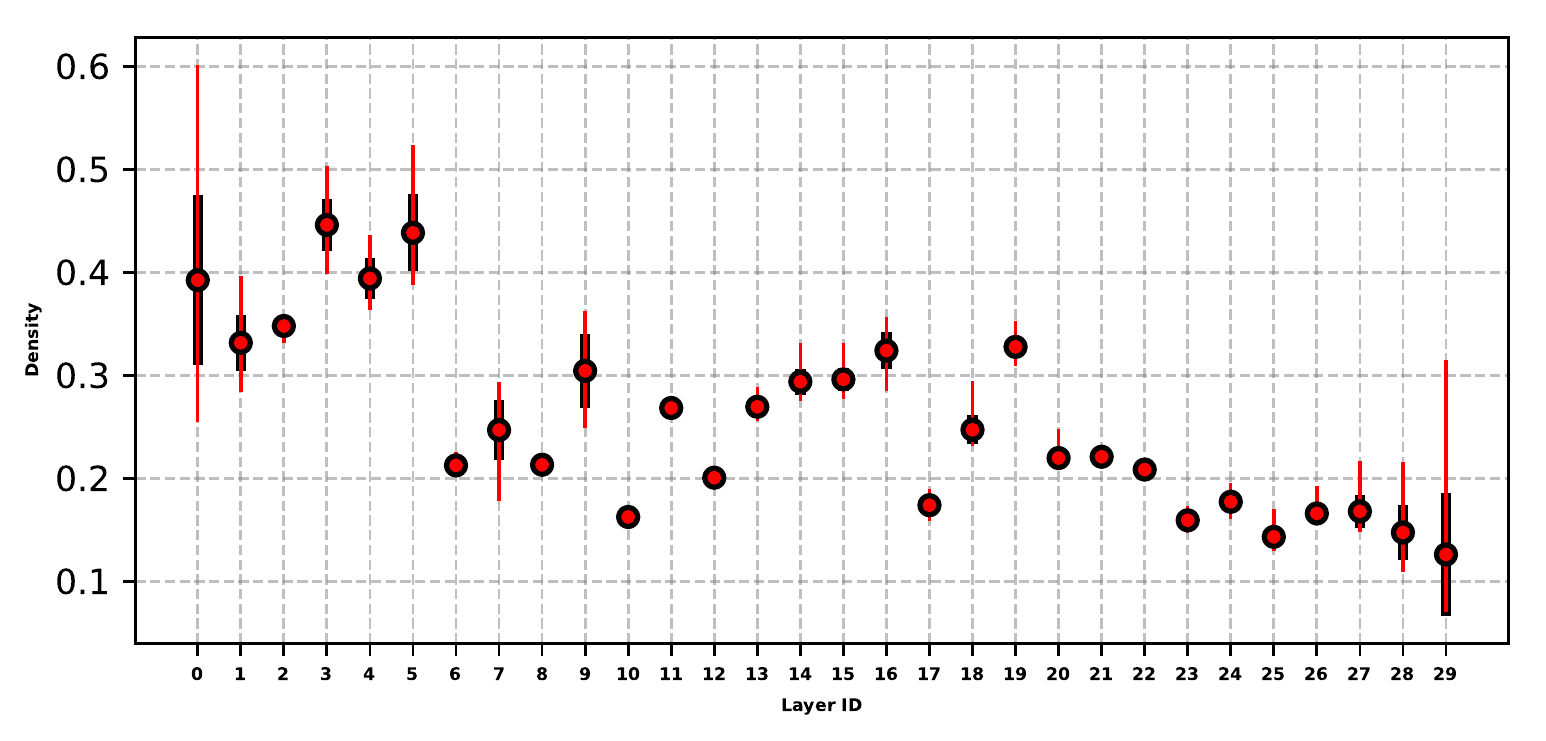}} 
    % \subfloat[Speed IV3 \label{fig:IV1}][]{\includegraphics[width=0.85\textheight]{figures/TimeSaving/IV3-Time-Saving.pdf}} \quad
    % \subfloat[Memory IV3 \label{fig:IV1}][]{\includegraphics[width=0.85\textheight]{figures/CR/IV3-CR.pdf}}
  \end{minipage}
%   \vspace{-3em}
  \caption{Per-layer Stationary Density in \textit{ResNet-V2-101} for 20 Images}
  \label{fig:density_ResNet_101}
\end{sidewaysfigure}

\begin{sidewaysfigure}[]
  \begin{minipage}{\textwidth}
  % without a b c 
  \captionsetup[subfigure]{labelformat=empty}
  \centering
    \subfloat[Density IV3 \label{fig:IV1}][]{\includegraphics[width=1.1\textheight]{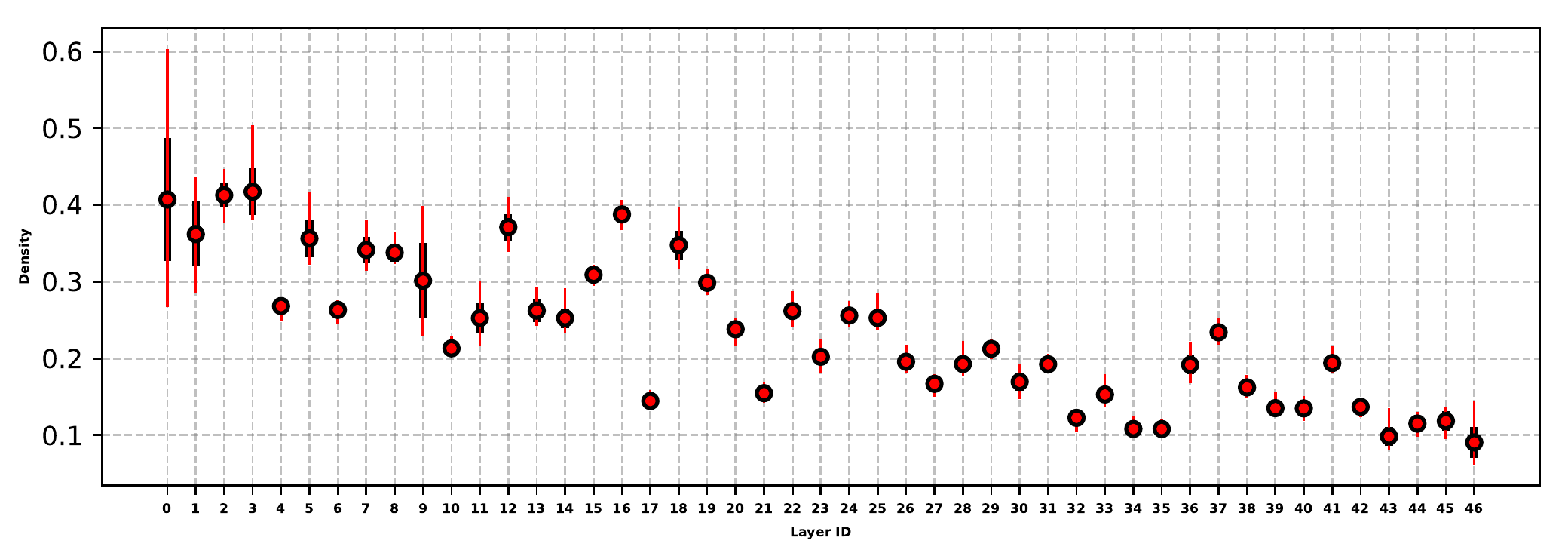}} 
    % \subfloat[Speed IV3 \label{fig:IV1}][]{\includegraphics[width=0.85\textheight]{figures/TimeSaving/IV3-Time-Saving.pdf}} \quad
    % \subfloat[Memory IV3 \label{fig:IV1}][]{\includegraphics[width=0.85\textheight]{figures/CR/IV3-CR.pdf}}
  \end{minipage}
%   \vspace{-3em}
  \caption{Per-layer Stationary Density in \textit{ResNet-V2-152} for 20 images}
  \label{fig:density_ResNet_152}
\end{sidewaysfigure}

\begin{figure*}[htbp]
  \begin{minipage}{\textwidth}
  % without a b c 
  \captionsetup[subfigure]{labelformat=empty}
  \centering
    \subfloat[Density IV1 \label{fig:IV1}][]{\includegraphics[width=1\textwidth,height=0.4\textwidth]{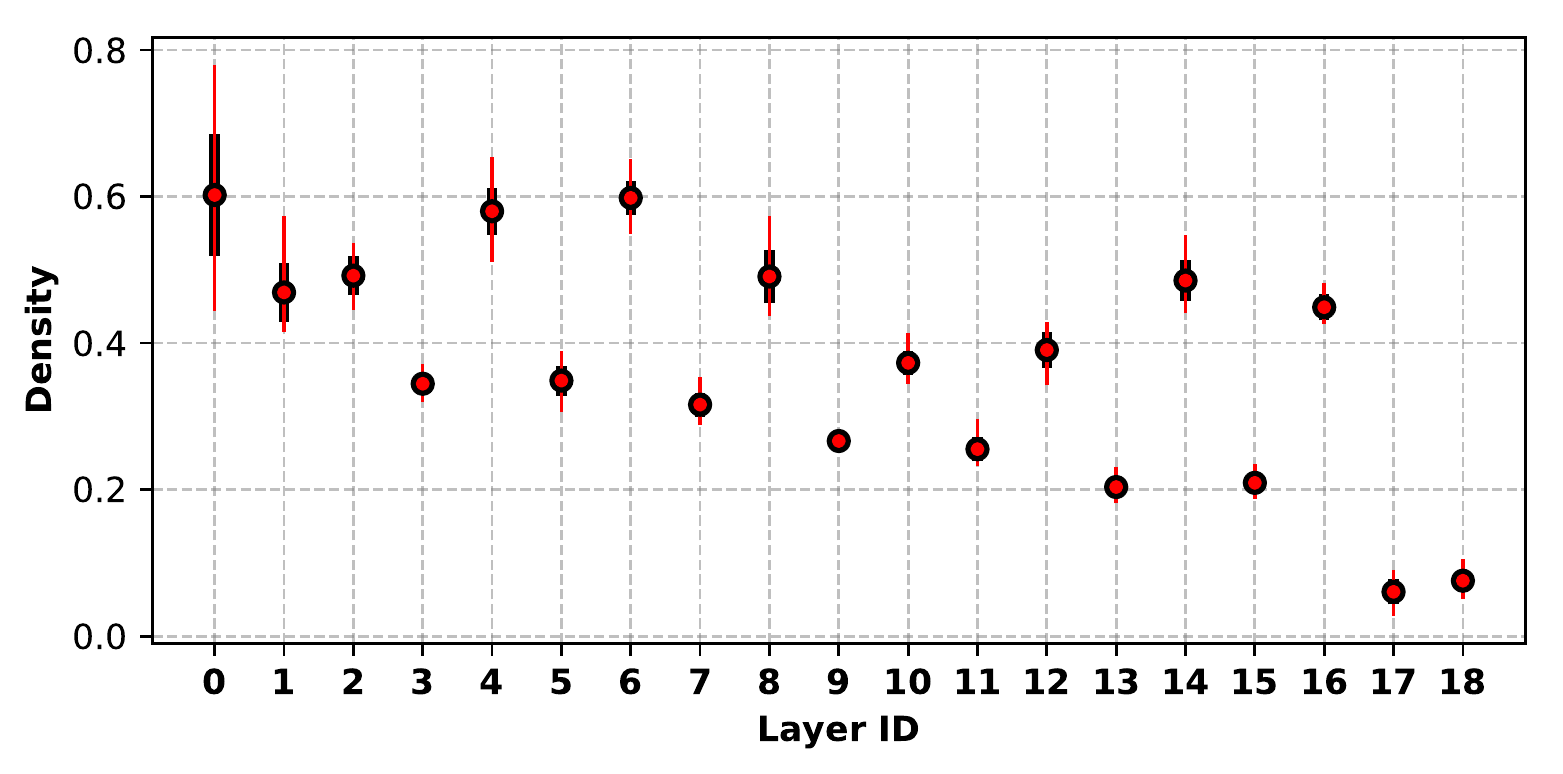}} 
    \\
    \subfloat[Speed IV1 \label{fig:IV1}][]{\includegraphics[width=1\textwidth,height=0.4\textwidth]{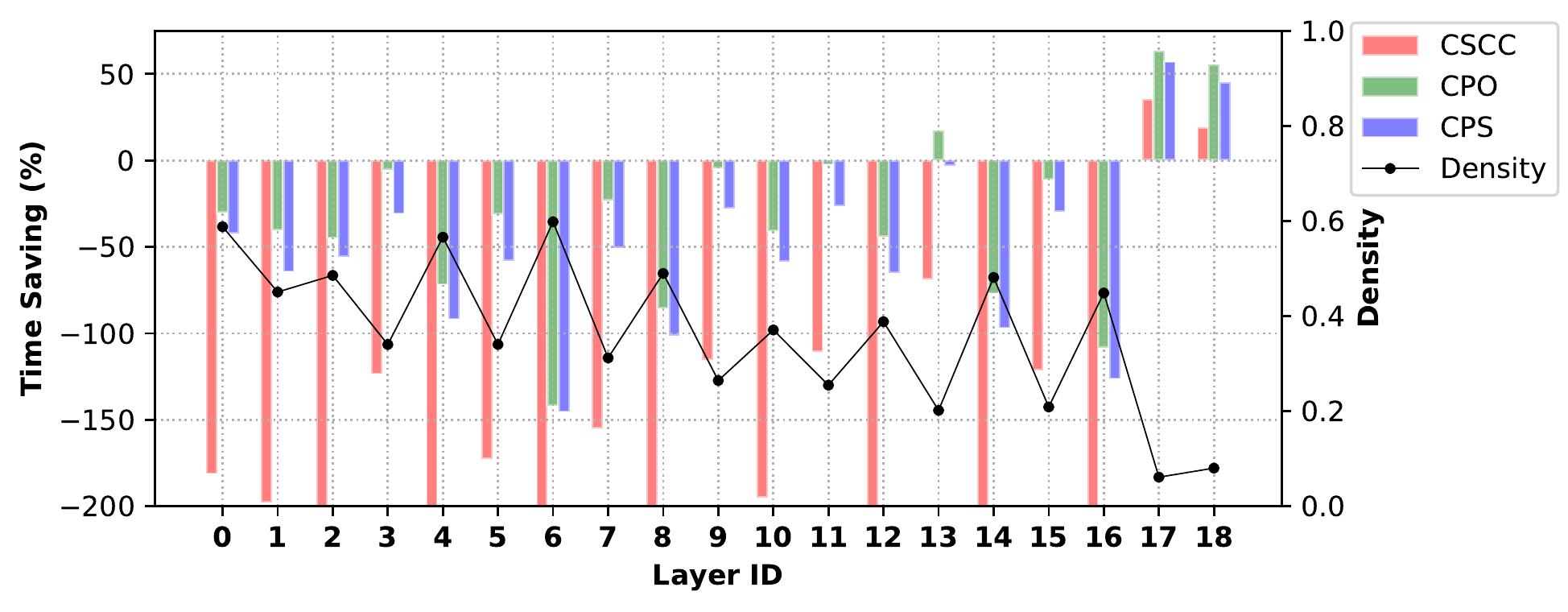}} \quad
    \subfloat[Memory IV1 \label{fig:IV1}][]{\includegraphics[width=1\textwidth,height=0.4\textwidth]{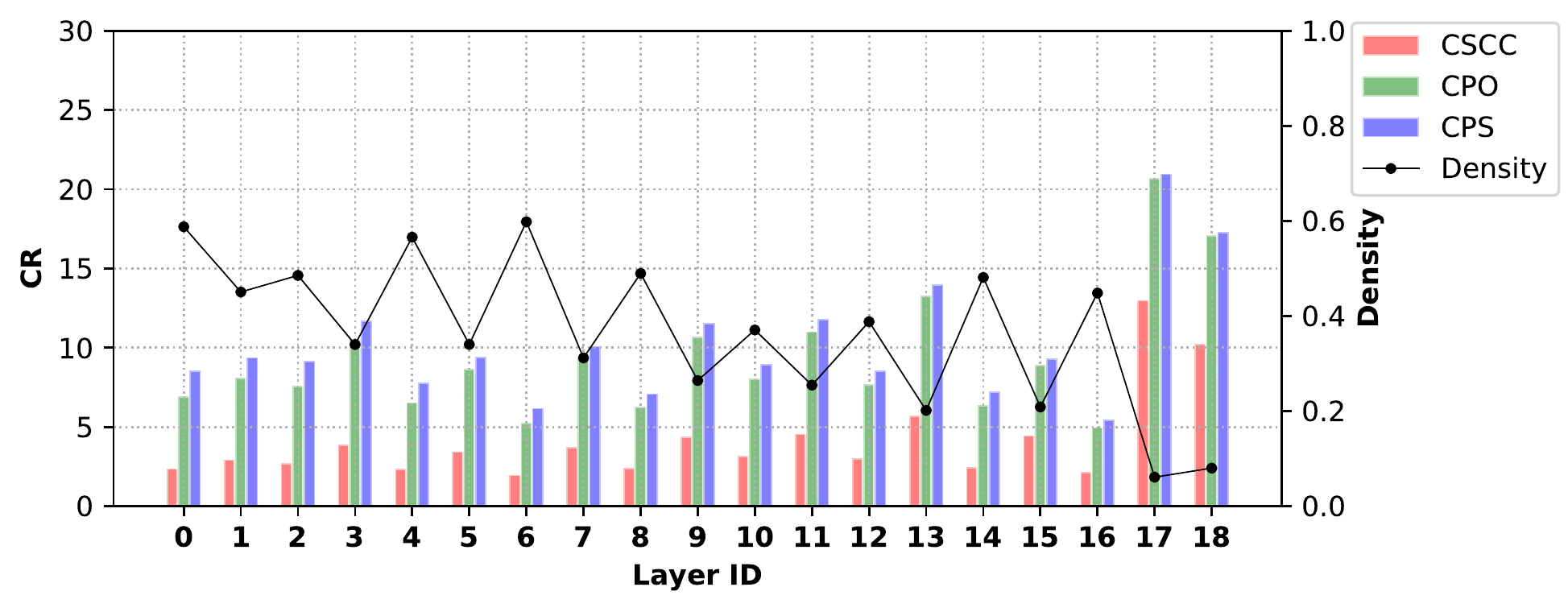}}
  \end{minipage}
%   \vspace{-3em}
  \caption{Top to bottom: Per-layer Stationary Density in \textit{IV1} for 20 Images, Per-layer Time Saving Results of CPO, CPS, and CSCC  in \textit{IV1} for 100 Images, Per-layer CR Results of CPO, CPS, and CSCC  in \textit{IV1} for 100 Images}
  \label{fig:density_speed_space_IV1}
\end{figure*}

\begin{sidewaysfigure}[]
  \begin{minipage}{\textwidth}
  % without a b c 
  \captionsetup[subfigure]{labelformat=empty}
  \centering
    \subfloat[Density IV3 \label{fig:IV1}][]{\includegraphics[width=1.1\textheight]{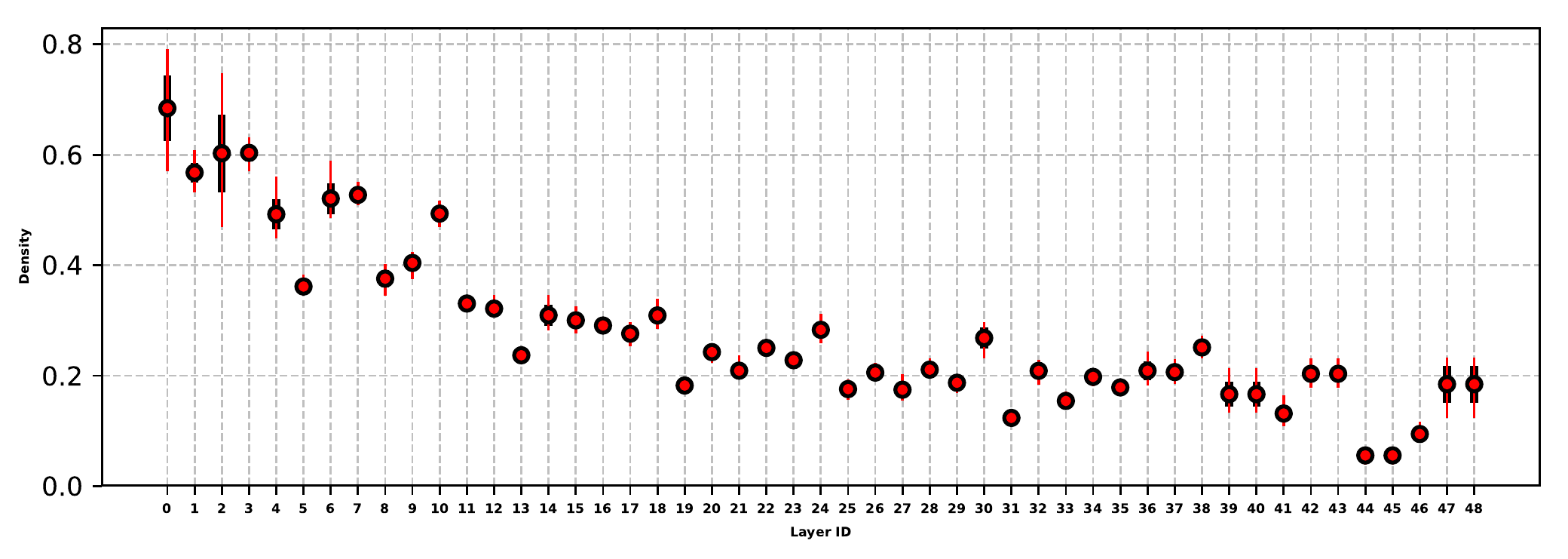}} 
    % \subfloat[Speed IV3 \label{fig:IV1}][]{\includegraphics[width=0.85\textheight]{figures/TimeSaving/IV3-Time-Saving.pdf}} \quad
    % \subfloat[Memory IV3 \label{fig:IV1}][]{\includegraphics[width=0.85\textheight]{figures/CR/IV3-CR.pdf}}
  \end{minipage}
%   \vspace{-3em}
  \caption{Per-layer Stationary Density in \textit{IV3} for 20 Images}
  \label{fig:density_IV3}
\end{sidewaysfigure}

\begin{sidewaysfigure}[]
%   \hspace{-9em}
  \begin{minipage}{\textwidth}
 
  % without a b c 
  \captionsetup[subfigure]{labelformat=empty}
  \centering
    \subfloat[Density IV3 \label{fig:IV1}][]{\includegraphics[width=1.1\textheight]{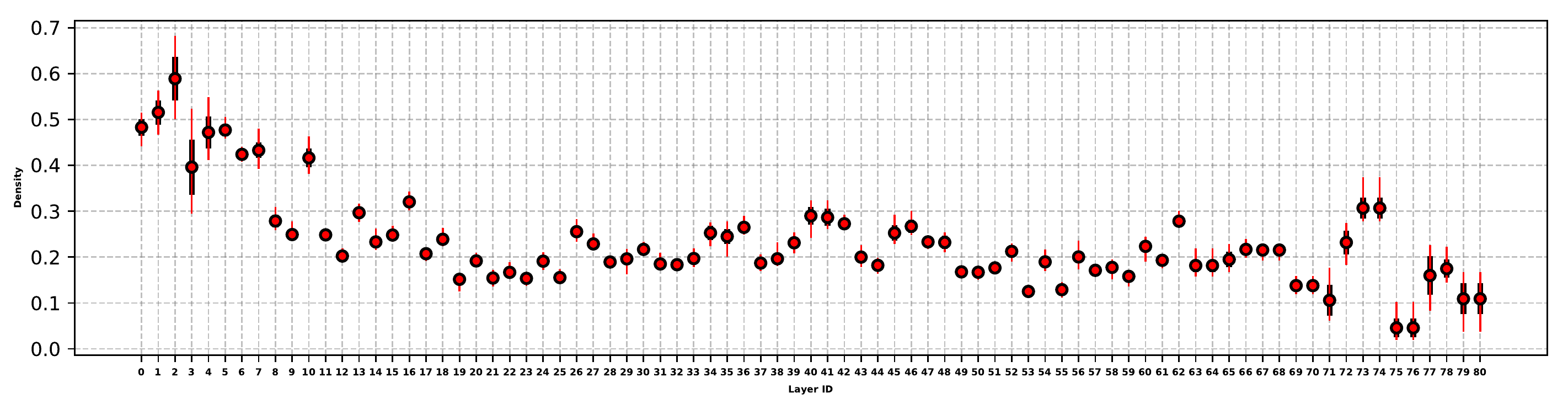}} 
    % \subfloat[Speed IV3 \label{fig:IV1}][]{\includegraphics[width=0.85\textheight]{figures/TimeSaving/IV3-Time-Saving.pdf}} \quad
    % \subfloat[Memory IV3 \label{fig:IV1}][]{\includegraphics[width=0.85\textheight]{figures/CR/IV3-CR.pdf}}
  \end{minipage}
%   \vspace{-3em}
  \caption{Per-layer Stationary Density in \textit{IV4} for 20 Images}
  \label{fig:density_IV4}
\end{sidewaysfigure}

In Figures \ref{fig:density_IV3}, \ref{fig:density_IV4}, \ref{fig:density_ResNet_101}, \ref{fig:density_ResNet_152}, the top of Figures \ref{fig:density_speed_space_ResNet_50}, \ref{fig:density_speed_space_IV1}, we are using a Boxplot to show the behaviour of density per layer through random 20 images. For each layer, we plot a line that consists of a midpoint, a bold line, and a thin line. At any given layer ID, the midpoint shows the mean of the density, the bold line shows the variance of the density, and the thin line shows
the density fluctuation of 20 images. As seen by the figures, the density in the majority of the layers is approximately stationary from one image to another. With this observation, the proposed hybrid selection approach was created to integrate the CPO/CPS into CNN inference.

\begin{sidewaysfigure}[]
  \begin{minipage}{\textwidth}
  % without a b c 
  \captionsetup[subfigure]{labelformat=empty}
  \centering
    % \subfloat[Density IV3 \label{fig:IV1}][]{\includegraphics[width=0.5\textheight]{figures/Density/ResNet-101-Density.pdf}} \quad
    \subfloat[Speed IV3 \label{fig:IV1}][]{\includegraphics[width=1\textheight]{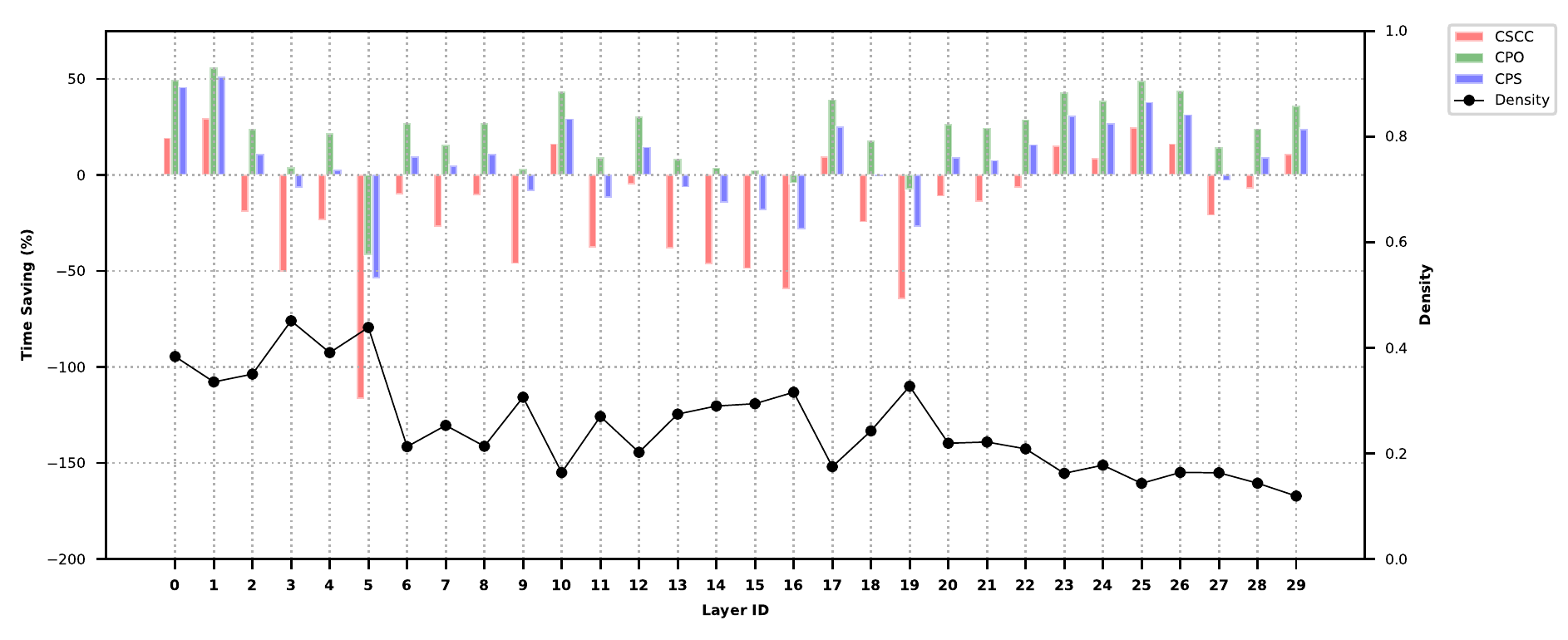}} \quad
    \subfloat[Memory IV3 \label{fig:IV1}][]{\includegraphics[width=1\textheight]{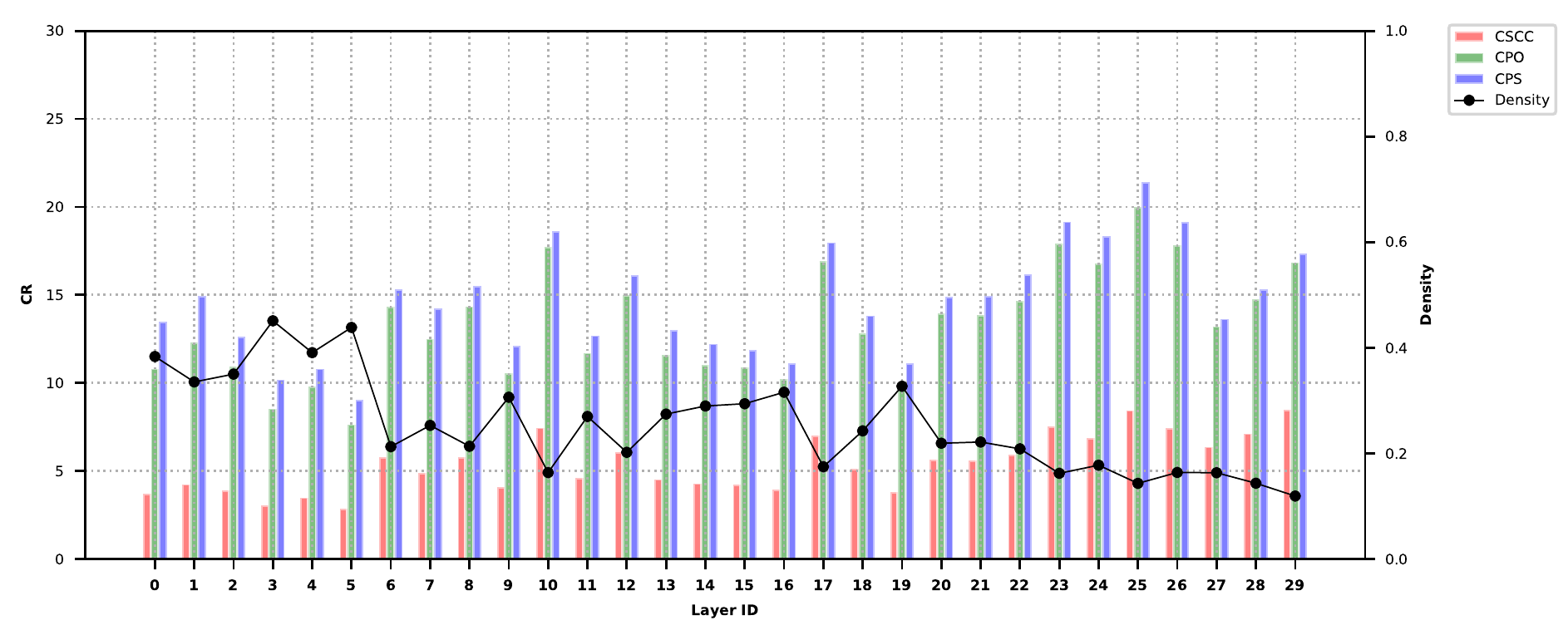}}
  \end{minipage}
%   \vspace{-3em}
  \caption{Per-layer Performance Results of CPO, CPS, and CSCC  in \textit{ResNet-V2-101} for 100 Images}
  \label{fig:speed_space_ResNet_101}
\end{sidewaysfigure}

\begin{sidewaysfigure}[]
\hspace{-9em}
  \begin{minipage}{\textwidth}
  % without a b c 
  \captionsetup[subfigure]{labelformat=empty}
  \centering
    % \subfloat[Density IV3 \label{fig:IV1}][]{\includegraphics[width=0.65\textheight]{figures/Density/IV3-Density.pdf}} \quad
    \subfloat[Speed IV3 \label{fig:IV1}][]{\includegraphics[width=1.2\textheight]{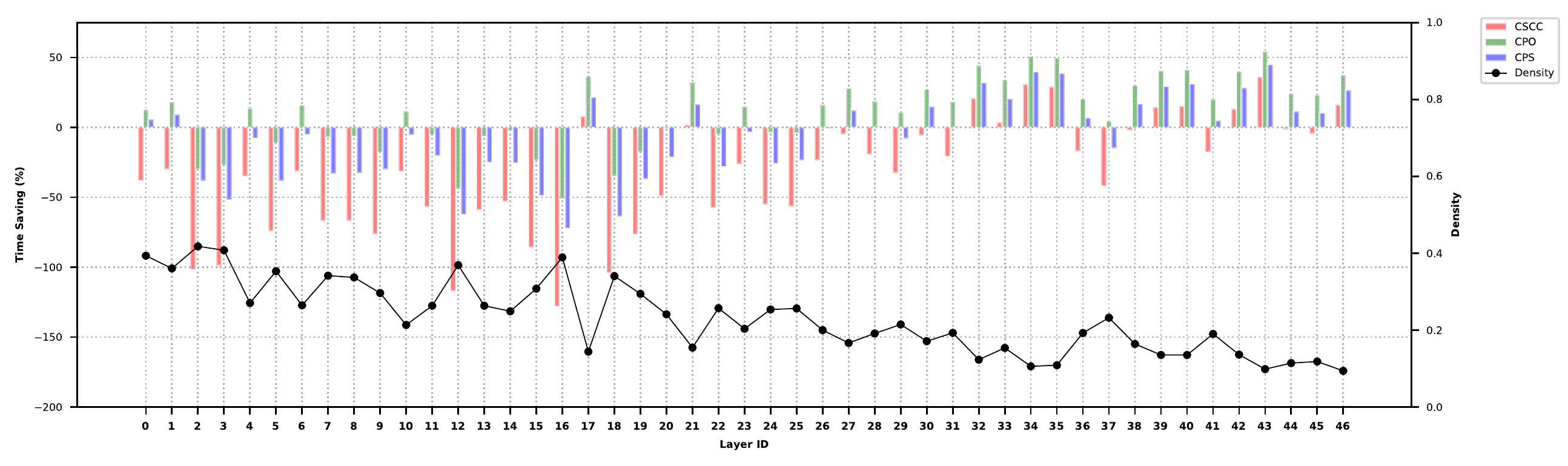}} \quad
    \subfloat[Memory IV3 \label{fig:IV1}][]{\includegraphics[width=1.2\textheight]{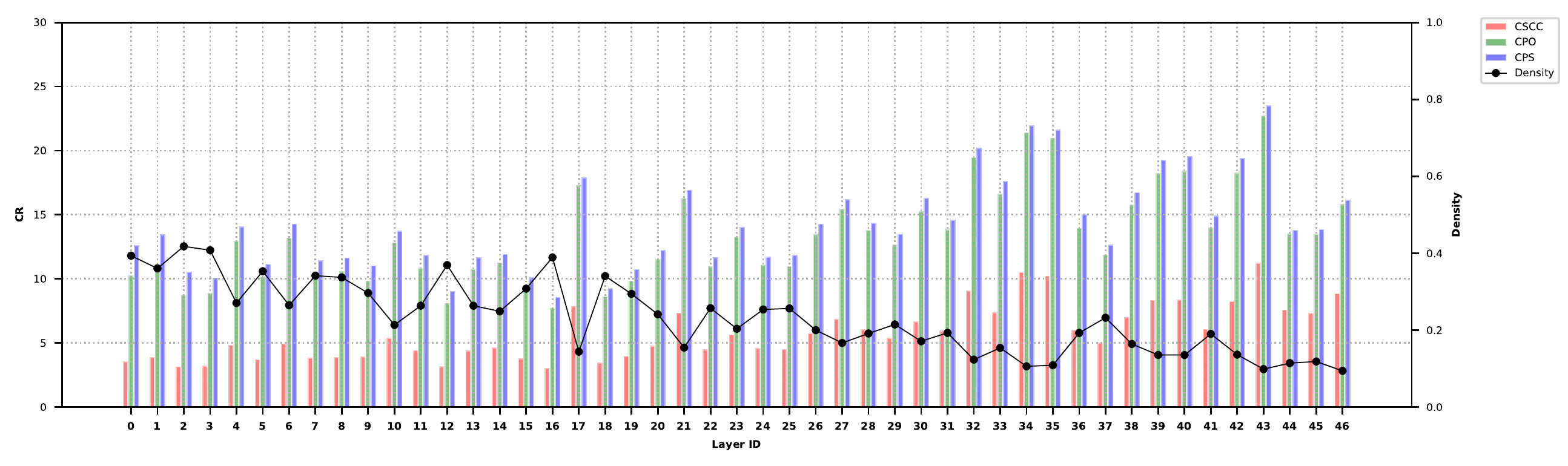}}
  \end{minipage}
%   \vspace{-3em}
  \caption{Per-layer Performance Results of CPO, CPS, and CSCC  in \textit{ResNet-V2-152} for 100 images}
  \label{fig:speed_space_ResNet_152}
\end{sidewaysfigure}

\begin{sidewaysfigure}[]
  \begin{minipage}{\textwidth}
  % without a b c 
  \captionsetup[subfigure]{labelformat=empty}
  \centering
    % \subfloat[Density IV3 \label{fig:IV1}][]{\includegraphics[width=0.65\textheight]{figures/Density/IV3-Density.pdf}} \quad
    \subfloat[Speed IV3 \label{fig:IV1}][]{\includegraphics[width=1.1\textheight]{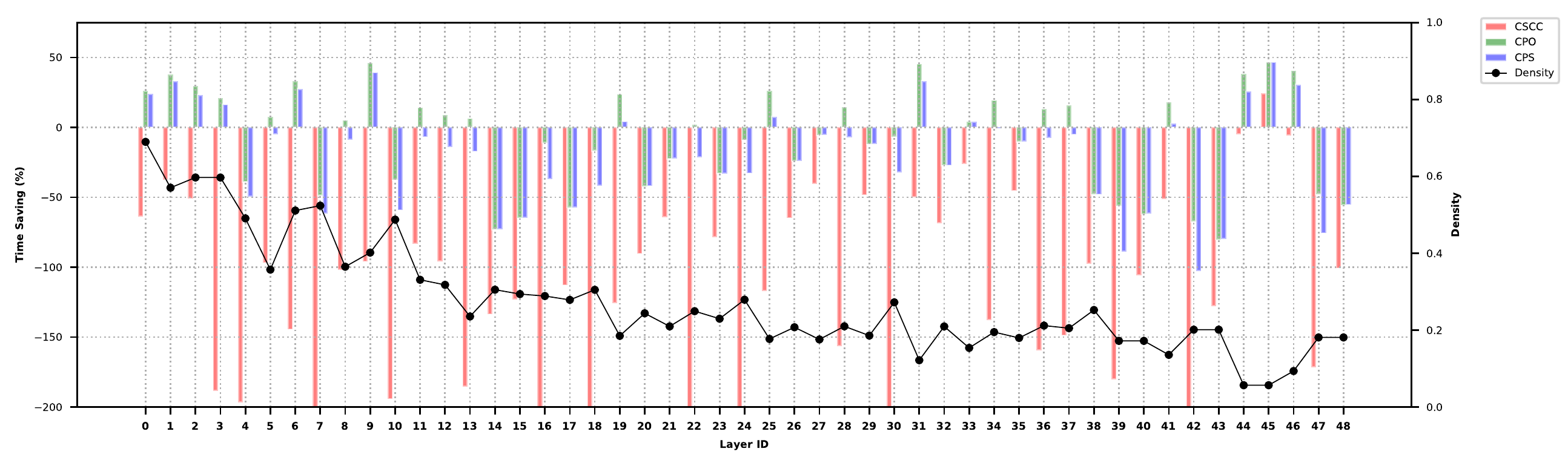}} \quad
    \subfloat[Memory IV3 \label{fig:IV1}][]{\includegraphics[width=1.1\textheight]{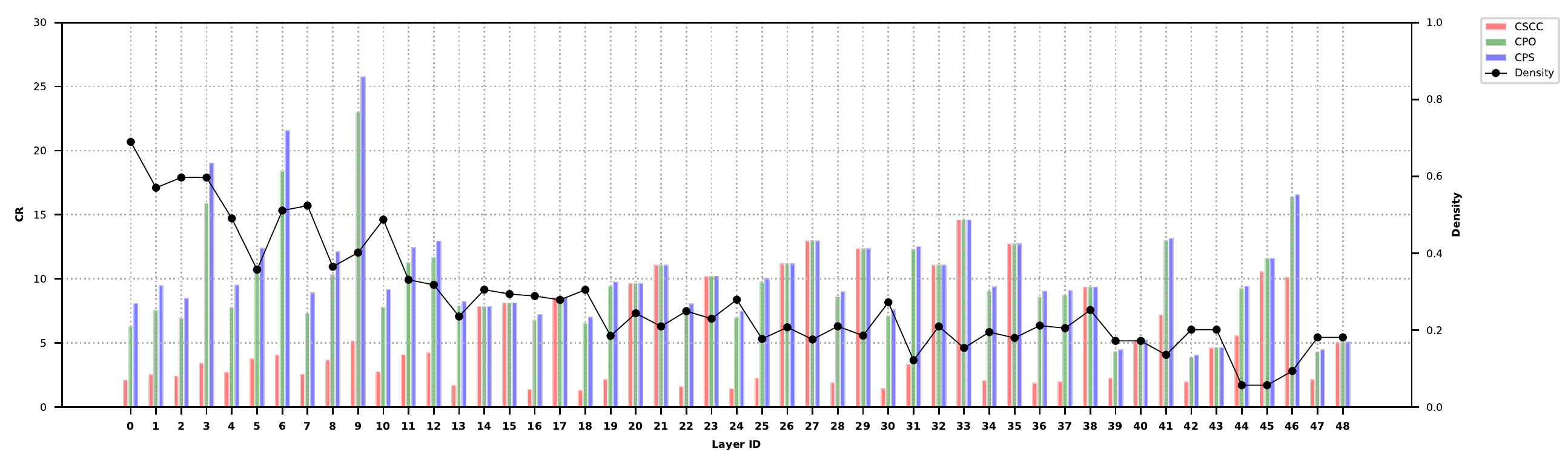}}
  \end{minipage}
%   \vspace{-3em}
  \caption{Per-layer Performance Results of CPO, CPS, and CSCC  in \textit{IV3} for 100 Images}
  \label{fig:speed_space_IV3}
\end{sidewaysfigure}

\begin{sidewaysfigure}[]
% \hspace{-9em}
  \begin{minipage}{\textwidth}
  % without a b c 
  \captionsetup[subfigure]{labelformat=empty}
  \centering
    % \subfloat[Density IV3 \label{fig:IV1}][]{\includegraphics[width=0.65\textheight]{figures/Density/IV3-Density.pdf}} \quad
    \subfloat[Speed IV3 \label{fig:IV1}][]{\includegraphics[width=1.1\textheight]{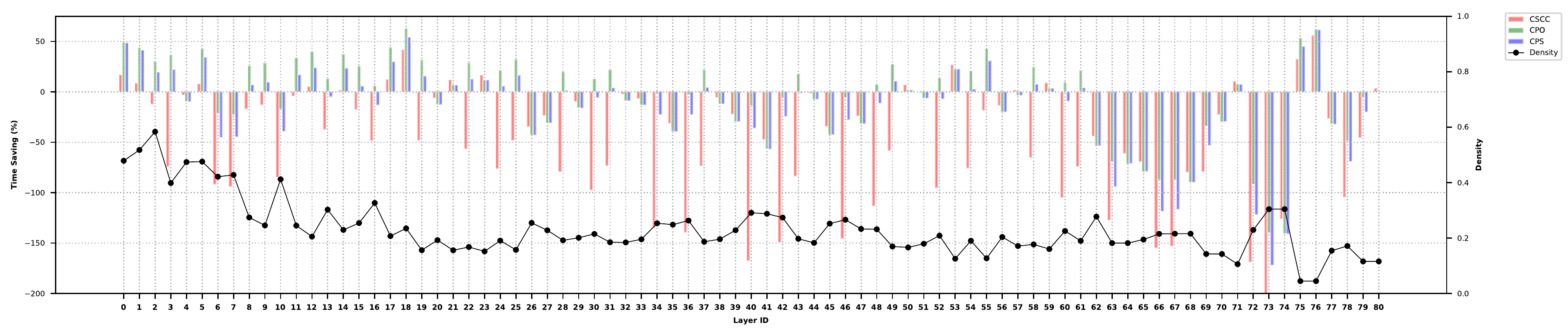}} \quad
    \subfloat[Memory IV3 \label{fig:IV1}][]{\includegraphics[width=1.1\linewidth]{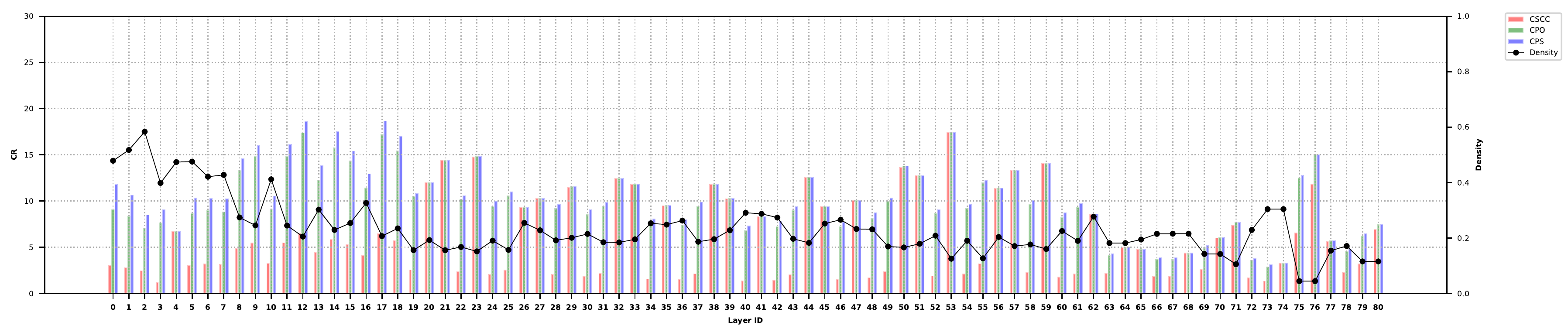}}
  \end{minipage}
%   \vspace{-3em}
  \caption{Per-layer Performance Results of CPO, CPS, and CSCC  in \textit{IV4} for 100 images}
  \label{fig:speed_space_IV4}
\end{sidewaysfigure}

This document presents two sets of figures for time and memory savings for each of CPO, CPS, and CSCC with respect to im2col: (1) At the middle of Figures \ref{fig:density_speed_space_ResNet_50}, \ref{fig:density_speed_space_IV1} and the top of Figures \ref{fig:speed_space_IV3}, \ref{fig:speed_space_IV4}, \ref{fig:speed_space_ResNet_101}, \ref{fig:speed_space_ResNet_152}, we show the per-layer time savings with respect to im2col of the layers already considered by CPO/CPS in all CNNs under test. (2) At the bottom of Figures \ref{fig:density_speed_space_ResNet_50}, \ref{fig:density_speed_space_IV1} and the bottom of Figures \ref{fig:speed_space_IV3}, \ref{fig:speed_space_IV4}, \ref{fig:speed_space_ResNet_101}, \ref{fig:speed_space_ResNet_152}, we show the compression ratio of the layers already considered by CPO/CPS in all CNNs under test. In each Figure, we plot a black line to show the per-layer density. As seen by these figures, CPO/CPS can achieve high time and memory savings in some layers with respect to im2col as already shown in the paper. In addition, we can observe that CPO/CPS is overall better than CSCC as already shown in the paper.

% %%%%%%%%%%%%

\subsection{ResNet-V2-50}

\begin{center}
\begin{longtable}{|l|l|l|l|l|l|l|l|l|l|l|l|l|l|}
\caption{ResNet-V2-50 DNN Analysis.} \label{tab:long4} \\

\hline \multicolumn{1}{|c|}{\textbf{ID}} & \multicolumn{1}{c|}{\textbf{Ih}} & \multicolumn{1}{c|}{\textbf{Iw}} & \multicolumn{1}{c|}{\textbf{Oh}} & \multicolumn{1}{c|}{\textbf{Ow}} &
\multicolumn{1}{c|}{\textbf{Kh}} & 
\multicolumn{1}{c|}{\textbf{Kw}} & 
\multicolumn{1}{c|}{\textbf{Sh}} & \multicolumn{1}{c|}{\textbf{Sw}} &
\multicolumn{1}{c|}{\textbf{Ic}} & 
\multicolumn{1}{c|}{\textbf{K}} 
\endfirsthead

\multicolumn{12}{c}%
{{\bfseries \tablename\ \thetable{} -- continued from previous page}} \\
\hline \multicolumn{1}{|c|}{\textbf{ID}} &  \multicolumn{1}{c}{\textbf{Ih}} & \multicolumn{1}{c|}{\textbf{Iw}} & \multicolumn{1}{c|}{\textbf{Oh}} & \multicolumn{1}{c|}{\textbf{Ow}} &
\multicolumn{1}{c|}{\textbf{Kh}} & 
\multicolumn{1}{c|}{\textbf{Kw}} & 
\multicolumn{1}{c|}{\textbf{Sh}} & \multicolumn{1}{c|}{\textbf{Sw}} &
\multicolumn{1}{c|}{\textbf{Ic}} & 
\multicolumn{1}{c|}{\textbf{K}} 
\\ \hline 
\endhead

\hline \multicolumn{12}{|r|}{{Continued on next page}} \\ \hline
\endfoot

\hline \hline % two hlines
\endlastfoot
\hline
0 & 75 & 75 & 75 & 75 & 3 & 3 & 1 & 1 & 64 & 64\\\hline
1 & 75 & 75 & 75 & 75 & 3 & 3 & 1 & 1 & 64 & 64\\\hline
2 & 38 & 38 & 38 & 38 & 3 & 3 & 1 & 1 & 128 & 128\\\hline
3 & 38 & 38 & 38 & 38 & 3 & 3 & 1 & 1 & 128 & 128\\\hline
4 & 38 & 38 & 38 & 38 & 3 & 3 & 1 & 1 & 128 & 128\\\hline
5 & 19 & 19 & 19 & 19 & 3 & 3 & 1 & 1 & 256 & 256\\\hline
6 & 19 & 19 & 19 & 19 & 3 & 3 & 1 & 1 & 256 & 256\\\hline
7 & 19 & 19 & 19 & 19 & 3 & 3 & 1 & 1 & 256 & 256\\\hline
8 & 19 & 19 & 19 & 19 & 3 & 3 & 1 & 1 & 256 & 256\\\hline
9 & 19 & 19 & 19 & 19 & 3 & 3 & 1 & 1 & 256 & 256\\\hline
10 & 10 & 10 & 10 & 10 & 3 & 3 & 1 & 1 & 512 & 512\\\hline
11 & 10 & 10 & 10 & 10 & 3 & 3 & 1 & 1 & 512 & 512\\\hline
12 & 10 & 10 & 10 & 10 & 3 & 3 & 1 & 1 & 512 & 512\\\hline
\end{longtable}
\end{center}

\newpage
% %%%%%%%%%%%%

% %%%%%%%%%%%%

\subsection{ResNet-V2-101}

\begin{center}
\begin{longtable}{|l|l|l|l|l|l|l|l|l|l|l|l|l|l|}
\caption{ResNet-V2-101 DNN Analysis.} \label{tab:long5} \\

\hline \multicolumn{1}{|c|}{\textbf{ID}} & \multicolumn{1}{c|}{\textbf{Ih}} & \multicolumn{1}{c|}{\textbf{Iw}} & \multicolumn{1}{c|}{\textbf{Oh}} & \multicolumn{1}{c|}{\textbf{Ow}} &
\multicolumn{1}{c|}{\textbf{Kh}} & 
\multicolumn{1}{c|}{\textbf{Kw}} & 
\multicolumn{1}{c|}{\textbf{Sh}} & \multicolumn{1}{c|}{\textbf{Sw}} &
\multicolumn{1}{c|}{\textbf{Ic}} & 
\multicolumn{1}{c|}{\textbf{K}} 
\endfirsthead

\multicolumn{12}{c}%
{{\bfseries \tablename\ \thetable{} -- continued from previous page}} \\
\hline \multicolumn{1}{|c|}{\textbf{ID}} &  \multicolumn{1}{c}{\textbf{Ih}} & \multicolumn{1}{c|}{\textbf{Iw}} & \multicolumn{1}{c|}{\textbf{Oh}} & \multicolumn{1}{c|}{\textbf{Ow}} &
\multicolumn{1}{c|}{\textbf{Kh}} & 
\multicolumn{1}{c|}{\textbf{Kw}} & 
\multicolumn{1}{c|}{\textbf{Sh}} & \multicolumn{1}{c|}{\textbf{Sw}} &
\multicolumn{1}{c|}{\textbf{Ic}} & 
\multicolumn{1}{c|}{\textbf{K}} 
\\ \hline 
\endhead

\hline \multicolumn{12}{|r|}{{Continued on next page}} \\ \hline
\endfoot

\hline \hline % two hlines
\endlastfoot
\hline
0 & 75 & 75 & 75 & 75 & 3 & 3 & 1 & 1 & 64 & 64\\\hline
1 & 75 & 75 & 75 & 75 & 3 & 3 & 1 & 1 & 64 & 64\\\hline
2 & 38 & 38 & 38 & 38 & 3 & 3 & 1 & 1 & 128 & 128\\\hline
3 & 38 & 38 & 38 & 38 & 3 & 3 & 1 & 1 & 128 & 128\\\hline
4 & 38 & 38 & 38 & 38 & 3 & 3 & 1 & 1 & 128 & 128\\\hline
5 & 19 & 19 & 19 & 19 & 3 & 3 & 1 & 1 & 256 & 256\\\hline
6 & 19 & 19 & 19 & 19 & 3 & 3 & 1 & 1 & 256 & 256\\\hline
7 & 19 & 19 & 19 & 19 & 3 & 3 & 1 & 1 & 256 & 256\\\hline
8 & 19 & 19 & 19 & 19 & 3 & 3 & 1 & 1 & 256 & 256\\\hline
9 & 19 & 19 & 19 & 19 & 3 & 3 & 1 & 1 & 256 & 256\\\hline
10 & 19 & 19 & 19 & 19 & 3 & 3 & 1 & 1 & 256 & 256\\\hline
11 & 19 & 19 & 19 & 19 & 3 & 3 & 1 & 1 & 256 & 256\\\hline
12 & 19 & 19 & 19 & 19 & 3 & 3 & 1 & 1 & 256 & 256\\\hline
13 & 19 & 19 & 19 & 19 & 3 & 3 & 1 & 1 & 256 & 256\\\hline
14 & 19 & 19 & 19 & 19 & 3 & 3 & 1 & 1 & 256 & 256\\\hline
15 & 19 & 19 & 19 & 19 & 3 & 3 & 1 & 1 & 256 & 256\\\hline
16 & 19 & 19 & 19 & 19 & 3 & 3 & 1 & 1 & 256 & 256\\\hline
17 & 19 & 19 & 19 & 19 & 3 & 3 & 1 & 1 & 256 & 256\\\hline
18 & 19 & 19 & 19 & 19 & 3 & 3 & 1 & 1 & 256 & 256\\\hline
19 & 19 & 19 & 19 & 19 & 3 & 3 & 1 & 1 & 256 & 256\\\hline
20 & 19 & 19 & 19 & 19 & 3 & 3 & 1 & 1 & 256 & 256\\\hline
21 & 19 & 19 & 19 & 19 & 3 & 3 & 1 & 1 & 256 & 256\\\hline
22 & 19 & 19 & 19 & 19 & 3 & 3 & 1 & 1 & 256 & 256\\\hline
23 & 19 & 19 & 19 & 19 & 3 & 3 & 1 & 1 & 256 & 256\\\hline
24 & 19 & 19 & 19 & 19 & 3 & 3 & 1 & 1 & 256 & 256\\\hline
25 & 19 & 19 & 19 & 19 & 3 & 3 & 1 & 1 & 256 & 256\\\hline
26 & 19 & 19 & 19 & 19 & 3 & 3 & 1 & 1 & 256 & 256\\\hline
27 & 10 & 10 & 10 & 10 & 3 & 3 & 1 & 1 & 512 & 512\\\hline
28 & 10 & 10 & 10 & 10 & 3 & 3 & 1 & 1 & 512 & 512\\\hline
29 & 10 & 10 & 10 & 10 & 3 & 3 & 1 & 1 & 512 & 512\\\hline
\end{longtable}
\end{center}

\newpage
%%%%%%%%%%%%

% %%%%%%%%%%%%

\subsection{ResNet-V2-152}

\begin{center}
\begin{longtable}{|l|l|l|l|l|l|l|l|l|l|l|l|l|l|}
\caption{ResNet-V2-152 DNN Analysis.} \label{tab:long6} \\

\hline \multicolumn{1}{|c|}{\textbf{ID}} & \multicolumn{1}{c|}{\textbf{Ih}} & \multicolumn{1}{c|}{\textbf{Iw}} & \multicolumn{1}{c|}{\textbf{Oh}} & \multicolumn{1}{c|}{\textbf{Ow}} &
\multicolumn{1}{c|}{\textbf{Kh}} & 
\multicolumn{1}{c|}{\textbf{Kw}} & 
\multicolumn{1}{c|}{\textbf{Sh}} & \multicolumn{1}{c|}{\textbf{Sw}} &
\multicolumn{1}{c|}{\textbf{Ic}} & 
\multicolumn{1}{c|}{\textbf{K}} 
\endfirsthead

\multicolumn{12}{c}%
{{\bfseries \tablename\ \thetable{} -- continued from previous page}} \\
\hline \multicolumn{1}{|c|}{\textbf{ID}} &  \multicolumn{1}{c}{\textbf{Ih}} & \multicolumn{1}{c|}{\textbf{Iw}} & \multicolumn{1}{c|}{\textbf{Oh}} & \multicolumn{1}{c|}{\textbf{Ow}} &
\multicolumn{1}{c|}{\textbf{Kh}} & 
\multicolumn{1}{c|}{\textbf{Kw}} & 
\multicolumn{1}{c|}{\textbf{Sh}} & \multicolumn{1}{c|}{\textbf{Sw}} &
\multicolumn{1}{c|}{\textbf{Ic}} & 
\multicolumn{1}{c|}{\textbf{K}} 
\\ \hline 
\endhead

\hline \multicolumn{12}{|r|}{{Continued on next page}} \\ \hline
\endfoot

\hline \hline % two hlines
\endlastfoot
\hline
0 & 56 & 56 & 56 & 56 & 3 & 3 & 1 & 1 & 64 & 64\\\hline
1 & 56 & 56 & 56 & 56 & 3 & 3 & 1 & 1 & 64 & 64\\\hline
2 & 28 & 28 & 28 & 28 & 3 & 3 & 1 & 1 & 128 & 128\\\hline
3 & 28 & 28 & 28 & 28 & 3 & 3 & 1 & 1 & 128 & 128\\\hline
4 & 28 & 28 & 28 & 28 & 3 & 3 & 1 & 1 & 128 & 128\\\hline
5 & 28 & 28 & 28 & 28 & 3 & 3 & 1 & 1 & 128 & 128\\\hline
6 & 28 & 28 & 28 & 28 & 3 & 3 & 1 & 1 & 128 & 128\\\hline
7 & 28 & 28 & 28 & 28 & 3 & 3 & 1 & 1 & 128 & 128\\\hline
8 & 28 & 28 & 28 & 28 & 3 & 3 & 1 & 1 & 128 & 128\\\hline
9 & 14 & 14 & 14 & 14 & 3 & 3 & 1 & 1 & 256 & 256\\\hline
10 & 14 & 14 & 14 & 14 & 3 & 3 & 1 & 1 & 256 & 256\\\hline
11 & 14 & 14 & 14 & 14 & 3 & 3 & 1 & 1 & 256 & 256\\\hline
12 & 14 & 14 & 14 & 14 & 3 & 3 & 1 & 1 & 256 & 256\\\hline
13 & 14 & 14 & 14 & 14 & 3 & 3 & 1 & 1 & 256 & 256\\\hline
14 & 14 & 14 & 14 & 14 & 3 & 3 & 1 & 1 & 256 & 256\\\hline
15 & 14 & 14 & 14 & 14 & 3 & 3 & 1 & 1 & 256 & 256\\\hline
16 & 14 & 14 & 14 & 14 & 3 & 3 & 1 & 1 & 256 & 256\\\hline
17 & 14 & 14 & 14 & 14 & 3 & 3 & 1 & 1 & 256 & 256\\\hline
18 & 14 & 14 & 14 & 14 & 3 & 3 & 1 & 1 & 256 & 256\\\hline
19 & 14 & 14 & 14 & 14 & 3 & 3 & 1 & 1 & 256 & 256\\\hline
20 & 14 & 14 & 14 & 14 & 3 & 3 & 1 & 1 & 256 & 256\\\hline
21 & 14 & 14 & 14 & 14 & 3 & 3 & 1 & 1 & 256 & 256\\\hline
22 & 14 & 14 & 14 & 14 & 3 & 3 & 1 & 1 & 256 & 256\\\hline
23 & 14 & 14 & 14 & 14 & 3 & 3 & 1 & 1 & 256 & 256\\\hline
24 & 14 & 14 & 14 & 14 & 3 & 3 & 1 & 1 & 256 & 256\\\hline
25 & 14 & 14 & 14 & 14 & 3 & 3 & 1 & 1 & 256 & 256\\\hline
26 & 14 & 14 & 14 & 14 & 3 & 3 & 1 & 1 & 256 & 256\\\hline
27 & 14 & 14 & 14 & 14 & 3 & 3 & 1 & 1 & 256 & 256\\\hline
28 & 14 & 14 & 14 & 14 & 3 & 3 & 1 & 1 & 256 & 256\\\hline
29 & 14 & 14 & 14 & 14 & 3 & 3 & 1 & 1 & 256 & 256\\\hline
30 & 14 & 14 & 14 & 14 & 3 & 3 & 1 & 1 & 256 & 256\\\hline
31 & 14 & 14 & 14 & 14 & 3 & 3 & 1 & 1 & 256 & 256\\\hline
32 & 14 & 14 & 14 & 14 & 3 & 3 & 1 & 1 & 256 & 256\\\hline
33 & 14 & 14 & 14 & 14 & 3 & 3 & 1 & 1 & 256 & 256\\\hline
34 & 14 & 14 & 14 & 14 & 3 & 3 & 1 & 1 & 256 & 256\\\hline
35 & 14 & 14 & 14 & 14 & 3 & 3 & 1 & 1 & 256 & 256\\\hline
36 & 14 & 14 & 14 & 14 & 3 & 3 & 1 & 1 & 256 & 256\\\hline
37 & 14 & 14 & 14 & 14 & 3 & 3 & 1 & 1 & 256 & 256\\\hline
38 & 14 & 14 & 14 & 14 & 3 & 3 & 1 & 1 & 256 & 256\\\hline
39 & 14 & 14 & 14 & 14 & 3 & 3 & 1 & 1 & 256 & 256\\\hline
40 & 14 & 14 & 14 & 14 & 3 & 3 & 1 & 1 & 256 & 256\\\hline
41 & 14 & 14 & 14 & 14 & 3 & 3 & 1 & 1 & 256 & 256\\\hline
42 & 14 & 14 & 14 & 14 & 3 & 3 & 1 & 1 & 256 & 256\\\hline
43 & 14 & 14 & 14 & 14 & 3 & 3 & 1 & 1 & 256 & 256\\\hline
44 & 7 & 7 & 7 & 7 & 3 & 3 & 1 & 1 & 512 & 512\\\hline
45 & 7 & 7 & 7 & 7 & 3 & 3 & 1 & 1 & 512 & 512\\\hline
46 & 7 & 7 & 7 & 7 & 3 & 3 & 1 & 1 & 512 & 512\\\hline

\end{longtable}
\end{center}

\newpage

%%%%%%%%%%%%%%%

% %%%%%%%%%%%%

\subsection{Inception V1 (IV1)}

\begin{center}
\begin{longtable}{|l|l|l|l|l|l|l|l|l|l|l|l|l|l|}
\caption{IV1 DNN Analysis.} \label{tab:long1} \\

\hline \multicolumn{1}{|c|}{\textbf{ID}} & \multicolumn{1}{c|}{\textbf{Ih}} & \multicolumn{1}{c|}{\textbf{Iw}} & \multicolumn{1}{c|}{\textbf{Oh}} & \multicolumn{1}{c|}{\textbf{Ow}} &
\multicolumn{1}{c|}{\textbf{Kh}} & 
\multicolumn{1}{c|}{\textbf{Kw}} & 
\multicolumn{1}{c|}{\textbf{Sh}} & \multicolumn{1}{c|}{\textbf{Sw}} &
\multicolumn{1}{c|}{\textbf{Ic}} & 
\multicolumn{1}{c|}{\textbf{K}} 
\endfirsthead

\multicolumn{12}{c}%
{{\bfseries \tablename\ \thetable{} -- continued from previous page}} \\
\hline \multicolumn{1}{|c|}{\textbf{ID}} &  \multicolumn{1}{c}{\textbf{Ih}} & \multicolumn{1}{c|}{\textbf{Iw}} & \multicolumn{1}{c|}{\textbf{Oh}} & \multicolumn{1}{c|}{\textbf{Ow}} &
\multicolumn{1}{c|}{\textbf{Kh}} & 
\multicolumn{1}{c|}{\textbf{Kw}} & 
\multicolumn{1}{c|}{\textbf{Sh}} & \multicolumn{1}{c|}{\textbf{Sw}} &
\multicolumn{1}{c|}{\textbf{Ic}} & 
\multicolumn{1}{c|}{\textbf{K}} 
\\ \hline 
\endhead

\hline \multicolumn{12}{|r|}{{Continued on next page}} \\ \hline
\endfoot

\hline \hline % two hlines
\endlastfoot
\hline
0 & 56 & 56 & 56 & 56 & 3 & 3 & 1 & 1 & 64 & 192\\\hline
1 & 28 & 28 & 28 & 28 & 3 & 3 & 1 & 1 & 96 & 128\\\hline
2 & 28 & 28 & 28 & 28 & 3 & 3 & 1 & 1 & 16 & 32\\\hline
3 & 28 & 28 & 28 & 28 & 3 & 3 & 1 & 1 & 128 & 192\\\hline
4 & 28 & 28 & 28 & 28 & 3 & 3 & 1 & 1 & 32 & 96\\\hline
5 & 14 & 14 & 14 & 14 & 3 & 3 & 1 & 1 & 96 & 208\\\hline
6 & 14 & 14 & 14 & 14 & 3 & 3 & 1 & 1 & 16 & 48\\\hline
7 & 14 & 14 & 14 & 14 & 3 & 3 & 1 & 1 & 112 & 224\\\hline
8 & 14 & 14 & 14 & 14 & 3 & 3 & 1 & 1 & 24 & 64\\\hline
9 & 14 & 14 & 14 & 14 & 3 & 3 & 1 & 1 & 128 & 256\\\hline
10 & 14 & 14 & 14 & 14 & 3 & 3 & 1 & 1 & 24 & 64\\\hline
11 & 14 & 14 & 14 & 14 & 3 & 3 & 1 & 1 & 144 & 288\\\hline
12 & 14 & 14 & 14 & 14 & 3 & 3 & 1 & 1 & 32 & 64\\\hline
13 & 14 & 14 & 14 & 14 & 3 & 3 & 1 & 1 & 160 & 320\\\hline
14 & 14 & 14 & 14 & 14 & 3 & 3 & 1 & 1 & 32 & 128\\\hline
15 & 7 & 7 & 7 & 7 & 3 & 3 & 1 & 1 & 160 & 320\\\hline
16 & 7 & 7 & 7 & 7 & 3 & 3 & 1 & 1 & 32 & 128\\\hline
17 & 7 & 7 & 7 & 7 & 3 & 3 & 1 & 1 & 192 & 384\\\hline
18 & 7 & 7 & 7 & 7 & 3 & 3 & 1 & 1 & 48 & 128\\\hline
\end{longtable}
\end{center}

\newpage

%%%%%%%%%%%%

\subsection{Inception V3 (IV3)}

\begin{center}
\begin{longtable}{|l|l|l|l|l|l|l|l|l|l|l|l|l|l|}
\caption{IV3 DNN Analysis.} \label{tab:long2} \\

\hline \multicolumn{1}{|c|}{\textbf{ID}} & \multicolumn{1}{c|}{\textbf{Ih}} & \multicolumn{1}{c|}{\textbf{Iw}} & \multicolumn{1}{c|}{\textbf{Oh}} & \multicolumn{1}{c|}{\textbf{Ow}} &
\multicolumn{1}{c|}{\textbf{Kh}} & 
\multicolumn{1}{c|}{\textbf{Kw}} & 
\multicolumn{1}{c|}{\textbf{Sh}} & \multicolumn{1}{c|}{\textbf{Sw}} &
\multicolumn{1}{c|}{\textbf{Ic}} & 
\multicolumn{1}{c|}{\textbf{K}} 
\endfirsthead

\multicolumn{12}{c}%
{{\bfseries \tablename\ \thetable{} -- continued from previous page}} \\
\hline \multicolumn{1}{|c|}{\textbf{ID}} &  \multicolumn{1}{c|}{\textbf{Ih}} & \multicolumn{1}{c|}{\textbf{Iw}} & \multicolumn{1}{c|}{\textbf{Oh}} & \multicolumn{1}{c|}{\textbf{Ow}} &
\multicolumn{1}{c|}{\textbf{Kh}} & 
\multicolumn{1}{c|}{\textbf{Kw}} & 
\multicolumn{1}{c|}{\textbf{sh}} & \multicolumn{1}{c|}{\textbf{sw}} &
\multicolumn{1}{c|}{\textbf{Ic}} & 
\multicolumn{1}{c|}{\textbf{K}} 
\\ \hline 
\endhead

\hline \multicolumn{12}{|r|}{{Continued on next page}} \\ \hline
\endfoot

\hline \hline % two hlines
\endlastfoot
\hline
0 & 149 & 149 & 147 & 147 & 3 & 3 & 1 & 1 & 32 & 32\\\hline
1 & 147 & 147 & 147 & 147 & 3 & 3 & 1 & 1 & 32 & 64\\\hline
2 & 73 & 73 & 71 & 71 & 3 & 3 & 1 & 1 & 80 & 192\\\hline
3 & 35 & 35 & 35 & 35 & 5 & 5 & 1 & 1 & 48 & 64\\\hline
4 & 35 & 35 & 35 & 35 & 3 & 3 & 1 & 1 & 64 & 96\\\hline
5 & 35 & 35 & 35 & 35 & 3 & 3 & 1 & 1 & 96 & 96\\\hline
6 & 35 & 35 & 35 & 35 & 5 & 5 & 1 & 1 & 48 & 64\\\hline
7 & 35 & 35 & 35 & 35 & 3 & 3 & 1 & 1 & 64 & 96\\\hline
8 & 35 & 35 & 35 & 35 & 3 & 3 & 1 & 1 & 96 & 96\\\hline
9 & 35 & 35 & 35 & 35 & 5 & 5 & 1 & 1 & 48 & 64\\\hline
10 & 35 & 35 & 35 & 35 & 3 & 3 & 1 & 1 & 64 & 96\\\hline
11 & 35 & 35 & 35 & 35 & 3 & 3 & 1 & 1 & 96 & 96\\\hline
12 & 35 & 35 & 35 & 35 & 3 & 3 & 1 & 1 & 64 & 96\\\hline
13 & 17 & 17 & 17 & 17 & 1 & 7 & 1 & 1 & 128 & 128\\\hline
14 & 17 & 17 & 17 & 17 & 7 & 1 & 1 & 1 & 128 & 192\\\hline
15 & 17 & 17 & 17 & 17 & 7 & 1 & 1 & 1 & 128 & 128\\\hline
16 & 17 & 17 & 17 & 17 & 1 & 7 & 1 & 1 & 128 & 128\\\hline
17 & 17 & 17 & 17 & 17 & 7 & 1 & 1 & 1 & 128 & 128\\\hline
18 & 17 & 17 & 17 & 17 & 1 & 7 & 1 & 1 & 128 & 192\\\hline
19 & 17 & 17 & 17 & 17 & 1 & 7 & 1 & 1 & 160 & 160\\\hline
20 & 17 & 17 & 17 & 17 & 7 & 1 & 1 & 1 & 160 & 192\\\hline
21 & 17 & 17 & 17 & 17 & 7 & 1 & 1 & 1 & 160 & 160\\\hline
22 & 17 & 17 & 17 & 17 & 1 & 7 & 1 & 1 & 160 & 160\\\hline
23 & 17 & 17 & 17 & 17 & 7 & 1 & 1 & 1 & 160 & 160\\\hline
24 & 17 & 17 & 17 & 17 & 1 & 7 & 1 & 1 & 160 & 192\\\hline
25 & 17 & 17 & 17 & 17 & 1 & 7 & 1 & 1 & 160 & 160\\\hline
26 & 17 & 17 & 17 & 17 & 7 & 1 & 1 & 1 & 160 & 192\\\hline
27 & 17 & 17 & 17 & 17 & 7 & 1 & 1 & 1 & 160 & 160\\\hline
28 & 17 & 17 & 17 & 17 & 1 & 7 & 1 & 1 & 160 & 160\\\hline
29 & 17 & 17 & 17 & 17 & 7 & 1 & 1 & 1 & 160 & 160\\\hline
30 & 17 & 17 & 17 & 17 & 1 & 7 & 1 & 1 & 160 & 192\\\hline
31 & 17 & 17 & 17 & 17 & 1 & 7 & 1 & 1 & 192 & 192\\\hline
32 & 17 & 17 & 17 & 17 & 7 & 1 & 1 & 1 & 192 & 192\\\hline
33 & 17 & 17 & 17 & 17 & 7 & 1 & 1 & 1 & 192 & 192\\\hline
34 & 17 & 17 & 17 & 17 & 1 & 7 & 1 & 1 & 192 & 192\\\hline
35 & 17 & 17 & 17 & 17 & 7 & 1 & 1 & 1 & 192 & 192\\\hline
36 & 17 & 17 & 17 & 17 & 1 & 7 & 1 & 1 & 192 & 192\\\hline
37 & 17 & 17 & 17 & 17 & 1 & 7 & 1 & 1 & 192 & 192\\\hline
38 & 17 & 17 & 17 & 17 & 7 & 1 & 1 & 1 & 192 & 192\\\hline
39 & 8 & 8 & 8 & 8 & 1 & 3 & 1 & 1 & 384 & 384\\\hline
40 & 8 & 8 & 8 & 8 & 3 & 1 & 1 & 1 & 384 & 384\\\hline
41 & 8 & 8 & 8 & 8 & 3 & 3 & 1 & 1 & 448 & 384\\\hline
42 & 8 & 8 & 8 & 8 & 1 & 3 & 1 & 1 & 384 & 384\\\hline
43 & 8 & 8 & 8 & 8 & 3 & 1 & 1 & 1 & 384 & 384\\\hline
44 & 8 & 8 & 8 & 8 & 1 & 3 & 1 & 1 & 384 & 384\\\hline
45 & 8 & 8 & 8 & 8 & 3 & 1 & 1 & 1 & 384 & 384\\\hline
46 & 8 & 8 & 8 & 8 & 3 & 3 & 1 & 1 & 448 & 384\\\hline
47 & 8 & 8 & 8 & 8 & 1 & 3 & 1 & 1 & 384 & 384\\\hline
48 & 8 & 8 & 8 & 8 & 3 & 1 & 1 & 1 & 384 & 384\\\hline
\end{longtable}
\end{center}

\newpage

% %%%%%%%%%%%%

\subsection{Inception V4 (IV4)}

\begin{center}
\begin{longtable}{|l|l|l|l|l|l|l|l|l|l|l|l|l|l|}
\caption{IV4 DNN Analysis.} \label{tab:long3} \\

\hline \multicolumn{1}{|c|}{\textbf{ID}} & \multicolumn{1}{c|}{\textbf{Ih}} & \multicolumn{1}{c|}{\textbf{Iw}} & \multicolumn{1}{c|}{\textbf{Oh}} & \multicolumn{1}{c|}{\textbf{Ow}} &
\multicolumn{1}{c|}{\textbf{Kh}} & 
\multicolumn{1}{c|}{\textbf{Kw}} & 
\multicolumn{1}{c|}{\textbf{Sh}} & \multicolumn{1}{c|}{\textbf{Sw}} &
\multicolumn{1}{c|}{\textbf{Ic}} & 
\multicolumn{1}{c|}{\textbf{K}} 
\endfirsthead

\multicolumn{12}{c}%
{{\bfseries \tablename\ \thetable{} -- continued from previous page}} \\
\hline \multicolumn{1}{|c|}{\textbf{ID}} &  \multicolumn{1}{c}{\textbf{Ih}} & \multicolumn{1}{c|}{\textbf{Iw}} & \multicolumn{1}{c|}{\textbf{Oh}} & \multicolumn{1}{c|}{\textbf{Ow}} &
\multicolumn{1}{c|}{\textbf{Kh}} & 
\multicolumn{1}{c|}{\textbf{Kw}} & 
\multicolumn{1}{c|}{\textbf{Sh}} & \multicolumn{1}{c|}{\textbf{Sw}} &
\multicolumn{1}{c|}{\textbf{Ic}} & 
\multicolumn{1}{c|}{\textbf{K}} 
\\ \hline 
\endhead

\hline \multicolumn{12}{|r|}{{Continued on next page}} \\ \hline
\endfoot

\hline \hline % two hlines
\endlastfoot
\hline
0 & 149 & 149 & 147 & 147 & 3 & 3 & 1 & 1 & 32 & 32\\\hline
1 & 147 & 147 & 147 & 147 & 3 & 3 & 1 & 1 & 32 & 64\\\hline
2 & 73 & 73 & 71 & 71 & 3 & 3 & 1 & 1 & 64 & 96\\\hline
3 & 73 & 73 & 73 & 73 & 1 & 7 & 1 & 1 & 64 & 64\\\hline
4 & 73 & 73 & 73 & 73 & 7 & 1 & 1 & 1 & 64 & 64\\\hline
5 & 73 & 73 & 71 & 71 & 3 & 3 & 1 & 1 & 64 & 96\\\hline
6 & 35 & 35 & 35 & 35 & 3 & 3 & 1 & 1 & 64 & 96\\\hline
7 & 35 & 35 & 35 & 35 & 3 & 3 & 1 & 1 & 64 & 96\\\hline
8 & 35 & 35 & 35 & 35 & 3 & 3 & 1 & 1 & 96 & 96\\\hline
9 & 35 & 35 & 35 & 35 & 3 & 3 & 1 & 1 & 64 & 96\\\hline
10 & 35 & 35 & 35 & 35 & 3 & 3 & 1 & 1 & 64 & 96\\\hline
11 & 35 & 35 & 35 & 35 & 3 & 3 & 1 & 1 & 96 & 96\\\hline
12 & 35 & 35 & 35 & 35 & 3 & 3 & 1 & 1 & 64 & 96\\\hline
13 & 35 & 35 & 35 & 35 & 3 & 3 & 1 & 1 & 64 & 96\\\hline
14 & 35 & 35 & 35 & 35 & 3 & 3 & 1 & 1 & 96 & 96\\\hline
15 & 35 & 35 & 35 & 35 & 3 & 3 & 1 & 1 & 64 & 96\\\hline
16 & 35 & 35 & 35 & 35 & 3 & 3 & 1 & 1 & 64 & 96\\\hline
17 & 35 & 35 & 35 & 35 & 3 & 3 & 1 & 1 & 96 & 96\\\hline
18 & 35 & 35 & 35 & 35 & 3 & 3 & 1 & 1 & 192 & 224\\\hline
19 & 17 & 17 & 17 & 17 & 1 & 7 & 1 & 1 & 192 & 224\\\hline
20 & 17 & 17 & 17 & 17 & 7 & 1 & 1 & 1 & 224 & 256\\\hline
21 & 17 & 17 & 17 & 17 & 7 & 1 & 1 & 1 & 192 & 192\\\hline
22 & 17 & 17 & 17 & 17 & 1 & 7 & 1 & 1 & 192 & 224\\\hline
23 & 17 & 17 & 17 & 17 & 7 & 1 & 1 & 1 & 224 & 224\\\hline
24 & 17 & 17 & 17 & 17 & 1 & 7 & 1 & 1 & 224 & 256\\\hline
25 & 17 & 17 & 17 & 17 & 1 & 7 & 1 & 1 & 192 & 224\\\hline
26 & 17 & 17 & 17 & 17 & 7 & 1 & 1 & 1 & 224 & 256\\\hline
27 & 17 & 17 & 17 & 17 & 7 & 1 & 1 & 1 & 192 & 192\\\hline
28 & 17 & 17 & 17 & 17 & 1 & 7 & 1 & 1 & 192 & 224\\\hline
29 & 17 & 17 & 17 & 17 & 7 & 1 & 1 & 1 & 224 & 224\\\hline
30 & 17 & 17 & 17 & 17 & 1 & 7 & 1 & 1 & 224 & 256\\\hline
31 & 17 & 17 & 17 & 17 & 1 & 7 & 1 & 1 & 192 & 224\\\hline
32 & 17 & 17 & 17 & 17 & 7 & 1 & 1 & 1 & 224 & 256\\\hline
33 & 17 & 17 & 17 & 17 & 7 & 1 & 1 & 1 & 192 & 192\\\hline
34 & 17 & 17 & 17 & 17 & 1 & 7 & 1 & 1 & 192 & 224\\\hline
35 & 17 & 17 & 17 & 17 & 7 & 1 & 1 & 1 & 224 & 224\\\hline
36 & 17 & 17 & 17 & 17 & 1 & 7 & 1 & 1 & 224 & 256\\\hline
37 & 17 & 17 & 17 & 17 & 1 & 7 & 1 & 1 & 192 & 224\\\hline
38 & 17 & 17 & 17 & 17 & 7 & 1 & 1 & 1 & 224 & 256\\\hline
39 & 17 & 17 & 17 & 17 & 7 & 1 & 1 & 1 & 192 & 192\\\hline
40 & 17 & 17 & 17 & 17 & 1 & 7 & 1 & 1 & 192 & 224\\\hline
41 & 17 & 17 & 17 & 17 & 7 & 1 & 1 & 1 & 224 & 224\\\hline
42 & 17 & 17 & 17 & 17 & 1 & 7 & 1 & 1 & 224 & 256\\\hline
43 & 17 & 17 & 17 & 17 & 1 & 7 & 1 & 1 & 192 & 224\\\hline
44 & 17 & 17 & 17 & 17 & 7 & 1 & 1 & 1 & 224 & 256\\\hline
45 & 17 & 17 & 17 & 17 & 7 & 1 & 1 & 1 & 192 & 192\\\hline
46 & 17 & 17 & 17 & 17 & 1 & 7 & 1 & 1 & 192 & 224\\\hline
47 & 17 & 17 & 17 & 17 & 7 & 1 & 1 & 1 & 224 & 224\\\hline
48 & 17 & 17 & 17 & 17 & 1 & 7 & 1 & 1 & 224 & 256\\\hline
49 & 17 & 17 & 17 & 17 & 1 & 7 & 1 & 1 & 192 & 224\\\hline
50 & 17 & 17 & 17 & 17 & 7 & 1 & 1 & 1 & 224 & 256\\\hline
51 & 17 & 17 & 17 & 17 & 7 & 1 & 1 & 1 & 192 & 192\\\hline
52 & 17 & 17 & 17 & 17 & 1 & 7 & 1 & 1 & 192 & 224\\\hline
53 & 17 & 17 & 17 & 17 & 7 & 1 & 1 & 1 & 224 & 224\\\hline
54 & 17 & 17 & 17 & 17 & 1 & 7 & 1 & 1 & 224 & 256\\\hline
55 & 17 & 17 & 17 & 17 & 1 & 7 & 1 & 1 & 192 & 224\\\hline
56 & 17 & 17 & 17 & 17 & 7 & 1 & 1 & 1 & 224 & 256\\\hline
57 & 17 & 17 & 17 & 17 & 7 & 1 & 1 & 1 & 192 & 192\\\hline
58 & 17 & 17 & 17 & 17 & 1 & 7 & 1 & 1 & 192 & 224\\\hline
59 & 17 & 17 & 17 & 17 & 7 & 1 & 1 & 1 & 224 & 224\\\hline
60 & 17 & 17 & 17 & 17 & 1 & 7 & 1 & 1 & 224 & 256\\\hline
61 & 17 & 17 & 17 & 17 & 1 & 7 & 1 & 1 & 256 & 256\\\hline
62 & 17 & 17 & 17 & 17 & 7 & 1 & 1 & 1 & 256 & 320\\\hline
63 & 8 & 8 & 8 & 8 & 1 & 3 & 1 & 1 & 384 & 256\\\hline
64 & 8 & 8 & 8 & 8 & 3 & 1 & 1 & 1 & 384 & 256\\\hline
65 & 8 & 8 & 8 & 8 & 3 & 1 & 1 & 1 & 384 & 448\\\hline
66 & 8 & 8 & 8 & 8 & 1 & 3 & 1 & 1 & 448 & 512\\\hline
67 & 8 & 8 & 8 & 8 & 1 & 3 & 1 & 1 & 512 & 256\\\hline
68 & 8 & 8 & 8 & 8 & 3 & 1 & 1 & 1 & 512 & 256\\\hline
69 & 8 & 8 & 8 & 8 & 1 & 3 & 1 & 1 & 384 & 256\\\hline
70 & 8 & 8 & 8 & 8 & 3 & 1 & 1 & 1 & 384 & 256\\\hline
71 & 8 & 8 & 8 & 8 & 3 & 1 & 1 & 1 & 384 & 448\\\hline
72 & 8 & 8 & 8 & 8 & 1 & 3 & 1 & 1 & 448 & 512\\\hline
73 & 8 & 8 & 8 & 8 & 1 & 3 & 1 & 1 & 512 & 256\\\hline
74 & 8 & 8 & 8 & 8 & 3 & 1 & 1 & 1 & 512 & 256\\\hline
75 & 8 & 8 & 8 & 8 & 1 & 3 & 1 & 1 & 384 & 256\\\hline
76 & 8 & 8 & 8 & 8 & 3 & 1 & 1 & 1 & 384 & 256\\\hline
77 & 8 & 8 & 8 & 8 & 3 & 1 & 1 & 1 & 384 & 448\\\hline
78 & 8 & 8 & 8 & 8 & 1 & 3 & 1 & 1 & 448 & 512\\\hline
79 & 8 & 8 & 8 & 8 & 1 & 3 & 1 & 1 & 512 & 256\\\hline
80 & 8 & 8 & 8 & 8 & 3 & 1 & 1 & 1 & 512 & 256\\\hline
\end{longtable}
\end{center}

\hspace{-6em}

\hspace{-1em}

\pagebreak
\section{CPO/CPS Encoding and Convolution Algorithms for K$_w$ = 1}
\label{sec:cpo_SI}

This subsection outlines the encoding and convolution algorithms for the proposed CPO for the special case K$_w$ = 1 (i.e, there are no overlap regions). This case is a straightforward extension of the algorithms presented in the paper (e.g,  Layer 10 in the paper).
Results for this case have already been presented in the paper, but we are providing the algorithms for this special case for clarity reasons. In addition, please note we chose to run CPO when K$_w$ = 1 between CPO and CPS because of more overall time saving. 
In Algorithm \ref{alg:cpoEncodeAlg_SI}, we simplify the complexity of the CSCC encoder in \cite{fan2019cscc} in terms of implementation as their algorithm is designed for any value of $K_w$ and $s_w$. Also, we added NPC in the CPO encoder to skip channels with all zeros. Algorithm \ref{alg:cpoConvAlg_SI} is mainly similar to the convolution algorithm  presented in \cite{fan2019cscc}, but with NPC support to skip the computations and storage of channels with all zeros.

\begin{algorithm}
    \caption{CPO/CPS Encoding Algorithm K$_w$ = 1}
    \label{alg:cpoEncodeAlg_SI}
    \begin{algorithmic}[1]
        \Procedure{cpoEncode}{}
            % \State $start = 0,$ $end = 8,$
            % \State Compute $\varepsilon_k$ from, where $start \leq k \leq end/2$
            %  \State Based on, $\forall_{j \in [end+1, end+4]}$ $\varepsilon_{j} = \varepsilon_{j-8},$
            %  \State $start = start + 4,$ 
            %  \State $end = end + 4,$ go to $\#3$
            \State \textbf{Input:} $ I$, $K_w$, $O_w$
            \State \textbf{Output:} $ptr$, $DA$, $IN$
            % \State $ptr\_sh=0$,\hspace{0.1em}$nz\_ch=0 $,\hspace{0.1em}$val\_sh=0$,\hspace{0.1em}$l=Padding_{Left}$ ,\hspace{0.1em}$i=0$ % initiate zero to all vairables
            % \State Initialize $ptr\_sh, nz\_ch, val\_sh, i$ to 0, pType to Pad\_Left
            \State Initialize ptr\_sh, val\_sh, s to 0
            \For{each channel \textbf{in} $I_{c}$} 
                \State nz\_ch, nz\_ptr, s = 0
                \State $ptr$[ptr\_sh++] = 0
                \For{$w$ = 0 to $I_w$ }
                    \If{$I(h,w) \neq 0$}
                        \State $DA$.insert($I(h,w)$)
                        \State $IN$.insert($w + (h * K_w) - s$)
                        \State nz\_ptr++
                    \EndIf
                    \If{$w$ =  s + $K_w$ - 1} \Comment{If End Submatrix}
                        \State $h$++
                        \State $w$ = s - 1
                        \If{$h$ = $I_h$}
                            \State $h$ = 0
                            \State s++
                            \State $w$ = s - 1
                            \State $ptr$[ptr\_sh++] = nz\_ptr
                            \If{$w$ = $I_w-K_w$}
                                \State Break Loop \Comment{End of Channel}
                            \EndIf
                        \EndIf
                    \EndIf
                \EndFor
                
                \If{nz\_ptr = 0} \Comment{If Skip Channel}
                    \State ptr\_sh -= ($O_w$ + 1)
                    \State $ptr$[ptr\_sh++] = NPC
                \EndIf
            \EndFor % Channel Loop
        \EndProcedure
    \end{algorithmic}
\end{algorithm}

% ------------------------------------------------------------------

\begin{algorithm}
    \caption{CPO/CPS Convolution Algorithm K$_w$ = 1}
    \label{alg:cpoConvAlg_SI}
    \begin{algorithmic}[1]
      \Function{convSpMv}{$K_h$, $K_w$, $O_h$, x, s, index}
            \For{$l = 0$ to $K_h$}
                \State y = (index / $K_w$) - l
                \State t = (index \% $K_w$ + $l*K_w$)
                \If{y $\ge$ 0 \hspace{0.1em} $\&$  y $<$ $O_h$}         \State O[y][s] += DA[x]*W[t]   
                  % the rest of the contributions
                %   \For{$i = 1$ to $pType$} 
                %      \State O[y][s+i] += DA[x]*W[t-i]
                %   \EndFor % i loop
              \EndIf % End if loop
             \EndFor % l loop
    \EndFunction
        \Procedure{cpoAlg}{}
        \State \textbf{Input:} $ptr$, $IN$, $DA$, $K_h$, $K_w$, $O_h$, $O_w$
        \State \textbf{Output:} $O$
            % \For{$c = 0$ to $I_{c}$}
            \State Initialize ptr\_sh = 0
            \For{each channel in $I_{c}$} 
            \IfThenOneLine {NPC \hspace{0.05em} \textrm{\textbf{then}}}% If ...2
            {$\text{Skip channel}$}% ...then...
            % % PNO Processing
            \State x = $ptr$[ptr\_sh] \Comment{First value of $ptr$ in each channel}
            
                % Submat loop
                \For{$s = 0$ to $O_w$}
                  \For{$tx = x$ to $ptr$[ptr\_sh + 1]}
                  
                    \State convSpMv($K_h$,$K_w$,$O_h$,$tx$, s,IN[tx])
                  \EndFor % x_ptr loop
                 
                 \State x = $ptr$[++ptr\_sh]
                \EndFor % submats loop
                
            \EndFor % end for channels
        \EndProcedure
    \end{algorithmic}
\end{algorithm}